\documentclass[letterpaper]{article} 
\def\year{2022}\relax
\usepackage{aaai22}  
\usepackage{times}  
\usepackage{helvet}  
\usepackage{courier}  
\usepackage[hyphens]{url}  
\usepackage{graphicx} 
\usepackage{natbib}  
\usepackage{caption} 
\DeclareCaptionStyle{ruled}{labelfont=normalfont,labelsep=colon,strut=off} 
\usepackage{algorithm}
\usepackage{algorithmic}
\usepackage{url}            
\usepackage{booktabs}       
\usepackage{amsfonts}       
\usepackage{nicefrac}       
\usepackage{microtype}      

\usepackage{xcolor}         

\usepackage{microtype}
\usepackage{graphicx}
\usepackage{subfigure}
\usepackage{booktabs} 

\usepackage{amsmath,amsfonts,bm}

\newcommand{\newterm}[1]{{\bf #1}}

\def\eqref#1{equation~\ref{#1}}

\def\1{\bm{1}}

\def\vtheta{{\bm{\theta}}}

\def\vp{{\bm{p}}}

\def\vx{{\bm{x}}}
\def\vy{{\bm{y}}}

\def\evy{{y}}

\DeclareMathAlphabet{\mathsfit}{\encodingdefault}{\sfdefault}{m}{sl}
\SetMathAlphabet{\mathsfit}{bold}{\encodingdefault}{\sfdefault}{bx}{n}

\def\sY{{\mathbb{Y}}}

\newcommand{\softmax}{\mathrm{softmax}}

\usepackage{amsmath}

\usepackage{varwidth}
\usepackage{caption}
\usepackage{wrapfig}
\usepackage{mathtools}
\usepackage{multirow}
\usepackage{nicefrac}
\usepackage{url}

\renewcommand{\paragraph}[1]{\textbf{#1}\hspace{1.8ex}}

\usepackage{newfloat}
\usepackage{listings}
\floatstyle{ruled}
\newfloat{listing}{tb}{lst}{}
\floatname{listing}{Listing}
\title{Well-Classified Examples are Underestimated in Classification with Deep Neural Networks}
\author{
    Guangxiang Zhao \textsuperscript{1, 2}, Wenkai Yang \textsuperscript{3}, Xuancheng Ren \textsuperscript{2}, Lei Li \textsuperscript{2}, Yunfang Wu \textsuperscript{2}, Xu Sun \textsuperscript{2, 4}\thanks{Corresponding Author}\\
}
\affiliations{

    \textsuperscript{1} Institute for Artificial Intelligence, Peking University\\
    \textsuperscript{2} MOE Key Laboratory of Computational Linguistics, School of Computer Science, Peking University\\
    \textsuperscript{3} Center for Data Science, Peking University \\
    \textsuperscript{4} Beijing Academy of Artificial Intelligence \\

    \{zhaoguangxiang, renxc, wuyf, xusun\}@pku.edu.cn, \{wkyang, lilei\}@stu.pku.edu.cn
}

\usepackage{bibentry}

\begin{document}

\maketitle

\begin{abstract}

The conventional wisdom behind learning deep classification models is to focus on bad-classified examples and ignore well-classified examples that are far from the decision boundary. For instance, when training with cross-entropy loss, examples with higher likelihoods (i.e., well-classified examples) contribute smaller gradients in back-propagation. However, we theoretically show that this common practice hinders representation learning, energy optimization, and margin growth. To counteract this deficiency, we propose to reward well-classified examples with additive bonuses to revive their contribution to the learning process. This counterexample theoretically addresses these three issues. We empirically support this claim by directly verifying the theoretical results or significant performance improvement with our counterexample on diverse tasks, including image classification, graph classification, and machine translation. Furthermore, this paper shows that we can deal with complex scenarios, such as imbalanced classification, OOD detection, and applications under adversarial attacks, because our idea can solve these three issues. Code is available at https://github.com/lancopku/well-classified-examples-are-underestimated.

\end{abstract}

\section{Introduction}

 In common practice, classification with deep neural networks~(DNNs) down-weights the contribution from well-classified examples. DNNs have achieved leading performance in mainstream classification tasks~\cite{resnet,kipf2016semi,transformer,bert}. Usually, the training of DNNs relies on optimizing the designed metrics between the target and the prediction through  back-propagation~\cite{mse-deep}. Mean-Square Error~(MSE) calculates a quadratic distance between the target and the probabilistic prediction of each example~\cite{mse-deep}. Cross-Entropy~(CE) loss measures the distance between the target distribution and the probability distribution~\cite{ce-deep}. CE loss is preferred as compared to MSE since CE loss encourages accurate predictions by bringing steep gradients to well-classified examples~\cite{ce-deep,why-ce}. Therefore, CE loss shows better generalization ability due to the enlarged steepness~\cite{back:ce-steep}. During training with CE loss, well-classified examples contribute less to the gradient as compared to bad-classified ones. 
 The wisdom behinds the operation of overlooking well-classified examples is well-classified examples have relatively less information in the learning. The improved variants of CE loss still comply with such wisdom~\cite{l-softmax,label-smoothing,focal_loss}.

\begin{figure}
\begin{center}
\centerline{\includegraphics[width=1.0\linewidth]{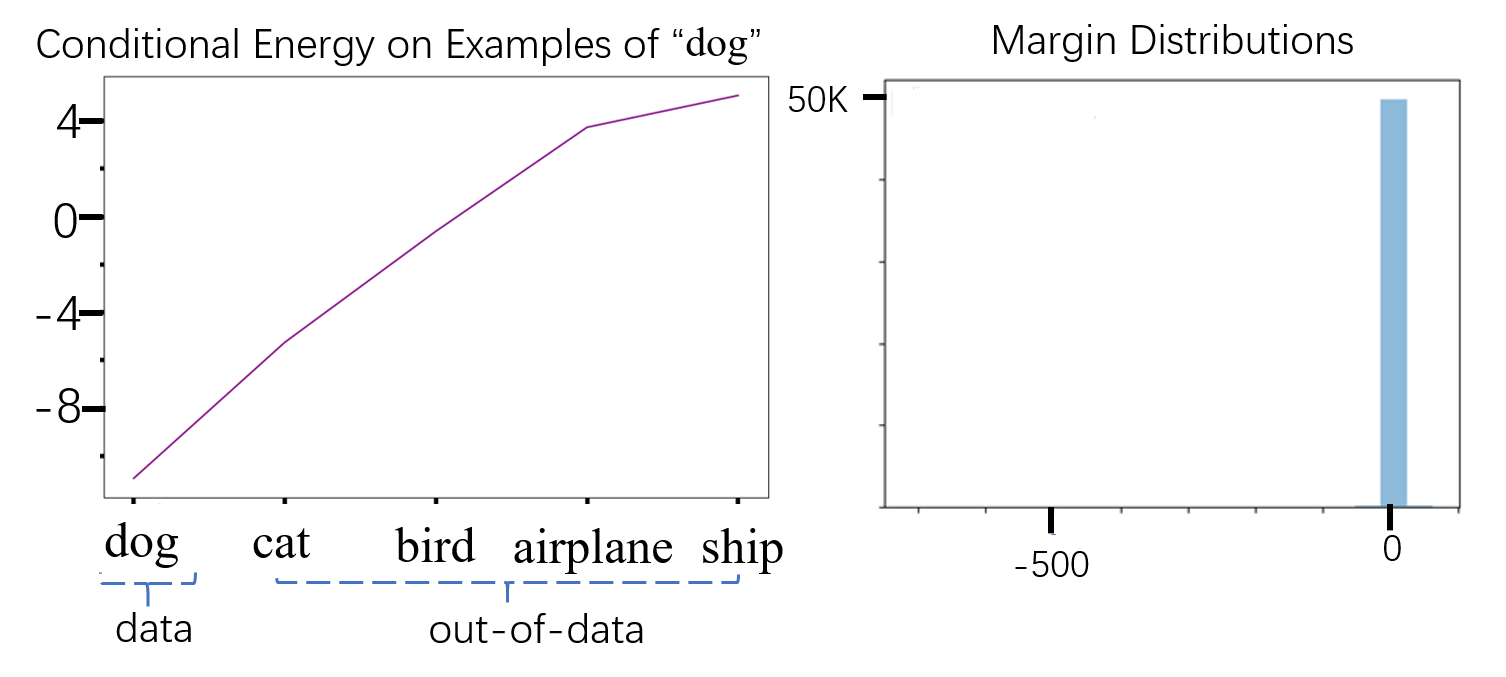}}
\caption{Illustration of energy and margins of CE loss on CIFAR-10. (Left): averaged conditional energy $E(y|x)$ for examples of class ``dog''. (Right): margin distributions of the trained classification models, there are some examples that have negative margins, most examples are around the decision boundary.}
\label{fig:energy-margin}
\end{center}
\end{figure}
We doubt the above practice with the following three facts: (1) Recent studies in imbalanced learning indicate that down-weighting the learning of relatively well-classified data-rich classes severely impairs the representation learning~\cite{Kang2020Decoupling,motivation:zhou2020bbn}. These studies inspire us to reflect on whether this is also the case at the sample level, and we validate that down-weighting the learning of well-classified samples also decreases the performance (Table~\ref{tab:imb}). (2) As to Energy-Based Models (EBM)~\cite{ebm}, a sharper energy surface is desired.\footnote{Refer to the talk ``The Future is Self-Supervised Learning'' by Yann LeCun at ICLR 2020} However, we find that the energy surface trained with CE loss is not sharp, as plotted in Figure \ref{fig:energy-margin}. The possible reason is that CE loss has insufficient power to push down the energy of the positive examples as long as it is lower than the energy of negative examples. Our validation in Figure~\ref{fig:energy} shows that up-weighting the well-classified examples returns a sharper surface. (3) As to classification, it is acknowledged that building classification models with large margins lead to good generalization~\cite{motivation:margin-generalization,motivation:margins} and good robustness~\cite{motivation:margin-robustness,motivation:margin-robustness-ijcnn,motivation:avg-margin-robustness}, but we find that the learning with CE loss leads to smaller margins (as plotted Figure \ref{fig:energy-margin}). The reason may be the incentive to further enlarge margins is limited, since well-classified examples are less optimized. Our results in Figure~\ref{fig:margin} and~\ref{fig:attack} show up-weighting classified samples enlarges the margin and helps to improve the adversarial robustness. 


\textbf{Contributions}: 
We systematically study the role of well-classified examples in classification learning with DNNs, and the results challenge the common practice. First, we theoretically identify issues of CE loss with back-propagation in terms of the learning of representations, the learning of energy functions, and the growth speed of margins (refer to \S~\ref{sec: CE loss}). Second, we propose Encouraging Loss~(EL), which can be viewed as a counterexample to the common practice since it up-weights the learning of well-classified examples compared with CE loss.  Besides, we theoretically demonstrate that by paying more attention to well-classified examples, learning with EL revives the representation learning from the part of well-classified examples, reduces the energy on the data manifold, and enlarges the classification margin (refer to \S~\ref{sec:method}). Third, we conduct extensive experiments to empirically verify the effectiveness of paying more attention to the well-classified instances for training canonical and state-of-the-art models. 
Results on image recognition and machine translation based on heterogeneous data types show that the counterexample can consistently improve learning performance. We also empirically verify that well-classified examples regarding representation learning, energy, and margins. 
Further analysis shows that enhancing the learning of well-classified examples improves models' capacity under complex scenarios, such as imbalanced classification, OOD detection, and application under adversarial attacks
(refer to \S~\ref{sec:practice}).
\section{Related Work}
Several studies are relevant to ours in three different aspects.




\paragraph{Representation learning}
There are several studies that direly or indirectly mitigate the shortage of representation learning from well-classified examples. Relying on various augmentation techniques, contrastive learning is one of the methods which directly mitigates the issue and has made outstanding progress in learning representations~\cite{unsupervised:moco,unsupervised:simclr,unsupervised:byol}. \citet{Kang2020Decoupling} mitigate the representation learning issue caused by class-level re-weighting, which indirectly down-weight well-classified examples, by turning it off in representation learning. However, they do not improve CE loss in representation learning.

\paragraph{Using energy-based models to interpret the classifier}
Recent studies re-interpret the classifier as EBM \cite{Grathwohl2020Your} or conditional EBM \cite{conditional-ebm1,conditional-ebm2} for generative models. However, our focus is to separately investigate the energy on the data and the data optimized by CE loss in classification models.

\paragraph{Enlarging the margin} 
A typical example of utilizing the idea of enlarging the minimum margin is the hinge loss \cite{suykens1999least} that is proposed to help learn a considerate margin in SVMs~\cite{cortes1995support} by focusing only on the data points close to the decision boundary.
Recently, the idea of enlarging margins has been introduced to the case of CE loss \cite{l-softmax,addtive-softmax,li2018angular,margin:logits,motivation:margin-robustness}.
Since these methods are based on logits adjustment before $\softmax$, they can be combined with our method by substituting the original CE loss with EL loss after $\softmax$. We show in the Appendix that our idea can be combined with their large margin idea.

\section{Exploring Theoretical Issues of CE Loss} \label{sec: CE loss}


\subsection{Setup and Notations}
\paragraph{Classification} In classification tasks, we have the input data $\vx \in \mathbb{R}^D$ and a label $y \in \mathbb{Y}$ which belongs to one of all $K$ classes. We use $\sY=\{1,2,...,K\}$ to denote the set of all class indices. We aim to learn a parametric function $f_{\vtheta}(\cdot)[\sY]$
that predicts $K$-dim logits (before normalization) for the data $\vx$, i.e., $f_{\vtheta}: \mathbb{R}^D \to \mathbb{R}^K$.
 Let $f_{\vtheta}(\vx)[\evy]$ denotes the $\evy$-th value of the predicted $K$-dim logits $f_{\vtheta}(\vx)[\sY]$, and $p_{\vtheta}(\vx)[\evy]$ is the normalized probability for the class $\evy$. 
 We adopt the general $\softmax$ normalization which transforms the logit $f(\vx)[\evy]$ to the  probability $p(\vx)[\evy]$ since it can generalize to both two-class and multi-class classification: $p_{\vtheta}(\vx)[\evy]=\softmax(f_{\vtheta}(\vx)[\evy]))=
     \frac{exp(f_{\vtheta}(\vx)[\evy])}{\sum_{\evy' \in \sY} exp(f_{\vtheta}(\vx)[\evy'])}$.
 
\paragraph{CE loss} Typically, the parametric classifier $f_{\vtheta}(\vx)[\sY]$ is estimated using Maximum Likelihood Estimation~(MLE), which is equivalent to minimizing the Cross-Entropy (CE) between the predicted probability $p(\vx)[\vy]$  and the signal that whether the class $\vy$ is the true label. When we get probabilities through the $\softmax$ normalization among all classes, we can simplify CE loss as minimizing negative log-likelihood (NLL) of $-\log p_{\theta}(\vx)[\evy]$, in which $\evy$ is the true label class for the input $\vx$.  The NLL loss is:
\begin{equation}\label{cf:nll}
\begin{aligned}
\mathcal{L}_{NLL}
&= -\log p_{\theta}(\evy\mid\vx) = - \log p_{\theta}(\vx)[\evy].
\end{aligned}
\end{equation}

In Eq. (\ref{cf:nll}), we get the predicted probability from the model $\vtheta$ and want to maximize the log probability of the target class $\evy$. We use the term \newterm{steepness of loss} to denote $\nicefrac{\partial \mathcal{L}}{\partial p}$.
For the NLL loss, the steepness of loss is $-\frac{1}{p}$, which means incorrect predicted examples with small $p$ are learned with priority, i.e., embodying a sharper loss and a larger steepness of loss. 
In this section, we refer CE loss to NLL loss and discuss gradients regarding NLL loss and $\softmax$ normalization. However, our results can also easily generalize to BCE loss with sigmoid since the opposite class of BCE loss in case of the second class in NLL loss, and the derivative of the sigmoid is the same as $\softmax$. In the NLL loss, we can only consider gradients from the index of the label class $\evy$. For simplicity, we use $p$ to denote the correctness of predictions.

\paragraph{Back-propagation}
\begin{equation} \label{cf:gradient}
\begin{aligned}
\frac{\partial \mathcal{L}}{\partial \vtheta} 
&= \sum_{y' \in \mathbb{Y}} \frac{\partial \mathcal{L}}{\partial {\sigma}_{\vtheta}(x)[y']}  \frac{\partial \sigma ( f_{\vtheta}(\vx)[y'])}{\partial \vtheta} \\ &= \sum_{y' \in \mathbb{Y}}  \frac{\partial \mathcal{L}}{\partial {\sigma}_{\vtheta}(x)[y']}  \frac{\partial \sigma ( f_{\vtheta}(\vx)[y'])}{\partial f_{\vtheta}(\vx)[y'] } \frac{\partial f_{\vtheta}(\vx)[y']}{\partial  \vtheta}.
\end{aligned}
\end{equation}
In Eq. (\ref{cf:gradient}), we can see that gradients depend on the loss function $\mathcal{L}$, logits normalization function $\sigma$  and the current model $f_{\vtheta}$. 
\subsection{Limitations of CE loss in Three Aspects}
\label{subsec: limitation of ce}
\paragraph{Normalization function brings a gradient vanishing problem to CE loss and hinders the representation learning}
At the beginning of the introduction of back-propagation to train deep neural networks, the loss function for back-propagation is MSE which measures the $L_{2}$ distance between probabilities and labels \cite{mse-deep}. However, the steepness of MSE gets to zero when the prediction gets close to the target. \citet{ce-deep} introduces CE loss to back-propagation and points out it addresses the above issue in MSE and makes predictions more accurate. Combining the derivative of NLL loss and the derivative of the normalization, gradients for the model parameters with the CE loss are:
\begin{equation} \label{cf:nll-gradient}
\begin{aligned}
\frac{\partial \mathcal{L}_{NLL}}{\partial \vtheta} 
=(p-1) \frac{\partial f_{\vtheta}(\vx)[y]}{\partial  \vtheta}.
\end{aligned}
\end{equation}
Thus, the normalization function brings the gradient saturation back as the prediction becomes correct. Since DNNs are regarded as a pipeline for extracting features~\cite{motivation:cnn-vis, motivation:bert-pipeline}, and it is in line with our perception that well-classified examples share pipeline with other examples, gradient vanishing hinders the part of representation learning from well-classified examples.


\paragraph{CE loss has insufficient power in reducing the energy on the data manifold}
Energy-based models (EBM) \cite{ebm} calculate the probability density of $p(x)$ for $x \in \mathbb{R}^D$ by: $p(x) =\frac{\exp(-E_{\vtheta}(x))}{\int_{x} \exp(-E_{\vtheta}(x))}$.
Here, we re-interpret the classifier as a conditional EBM: $p_{\theta}(y \mid \vx) = \frac{\exp{f_{\vtheta}(\vx)[y]}}{\sum_{y' \in \mathbb{Y}} \exp(f_{\vtheta}(\vx)[y'])}$, in which the conditional energy is $E_{\theta}(y \mid \vx)=-f_{\vtheta}(\vx)[y]$. Thus, the CE loss in Eq. (\ref{cf:nll}) can be written as:
\begin{equation}\label{cf:energy-ce}
\begin{aligned}
    \mathcal{L}_{NLL} 
    = E_{\theta}(y \mid \vx) &+ log[\exp(-E_{\vtheta}(y\mid\vx)) \\ &+ \sum_{y' \ne y} \exp(-E_{\vtheta}(y' \mid \vx))].
\end{aligned}
\end{equation}
Minimizing the CE loss can push up the energy out of the data $E_{\theta}(y'\mid\vx)$ where $y' \ne y$, but for the energy on the data manifold $E_{\theta}(y\mid\vx)$, although the nominator of the $\softmax$ function pulls that energy down, numerator which contains that term pushes the energy on the data up. Therefore, the learning with CE loss gets into a dilemma and has insufficient power to reduce the data's energy.

\paragraph{CE loss is not effective in enlarging margins}
Previous studies prove that large minimum margin~\cite{motivation:margin-generalization,NIPS2017_b22b257a,neyshabur2018a} or large overall margins~\cite{margin-distribution,motivation:margins} on the training set indicate good generalization ability. 
Though the margin $\gamma (\vx,y)=f_{\vtheta}(\vx)[\evy]- max_{y' \ne \evy}f_{\vtheta}(\vx)[y']$  is defined on the logits, since the $\softmax$ function generates probabilities against each other and has the exponential term that mimics the max operation, a larger likelihood is likely to lead to a larger margin. 
However, as we have deduced in Eq. (\ref{cf:nll-gradient}), CE loss is not good at increasing the likelihood when the likelihood becomes larger. Hence it is likely to be limited in increasing the margin.
Following previous work~\cite{addtive-softmax,LDAM,margin:logits}, we view the gap $f_{\vtheta}(\vx)[\evy]- f_{\vtheta}(\vx)[y']$ between the logit at the label position and the logit at any other position $y' \ne y$  as the approximated margin. 
Note that the NLL loss can then be written as $\mathcal{L}_{NLL} =log[1+\sum_{y' \ne y} \exp(f_{\vtheta}(\vx)[y'] -f_{\vtheta}(\vx)[y])]$. 

We use $A$ to denote $\exp(f_{\vtheta}(\vx)[y'] -f_{\vtheta}(\vx)[y])$, the gradients of the NLL loss w.r.t. the parameter $\vtheta$ is:
\begin{equation}\label{cf:ce-gradient-margin}
\begin{aligned}
\frac{\partial \mathcal{L}_{NLL}}{\partial \vtheta} =\frac{\sum_{y' \ne y} A (\frac{\partial ( f_{\vtheta}(\vx)[y']- f_{\vtheta}(\vx)[y])}{\partial  \vtheta})}{1+\sum_{y' \ne y} A} .
\end{aligned}
\end{equation}
The above formula interprets the training procedure of CE loss as increasing the logit for the label $f_{\vtheta}(\vx)[y]$, but decreasing the logits for other classes $f_{\vtheta}(\vx)[y']$. Therefore, it enlarges the gap, and the standard margin $f_{\vtheta}(\vx)[y]-\max_{y' \ne y}f_{\vtheta}(\vx)[y']$ is likely to be larger. However, when the prediction gets close to the target during training, $A$ gets close to $0$, while the numerator has a constant $1$, so the incentive for further enlarging the margin gets close to $0$. Therefore, CE loss is not effective in enlarging margins to a certain extent.

\section{Gaining from Reviving the Learning of Well-classified Examples} \label{sec:method}
 In this section, we propose a counterexample \textbf{Encouraging Loss} (EL) which increases the relative importance of well-classified examples in optimization compared to that in CE loss. We first define EL and then demonstrate that it can mitigate issues of CE loss mentioned before.

\subsection{The Counterexample: Encouraging Loss}
\begin{figure}[t]
\begin{center}
\centerline{\includegraphics[width=1.2\linewidth]{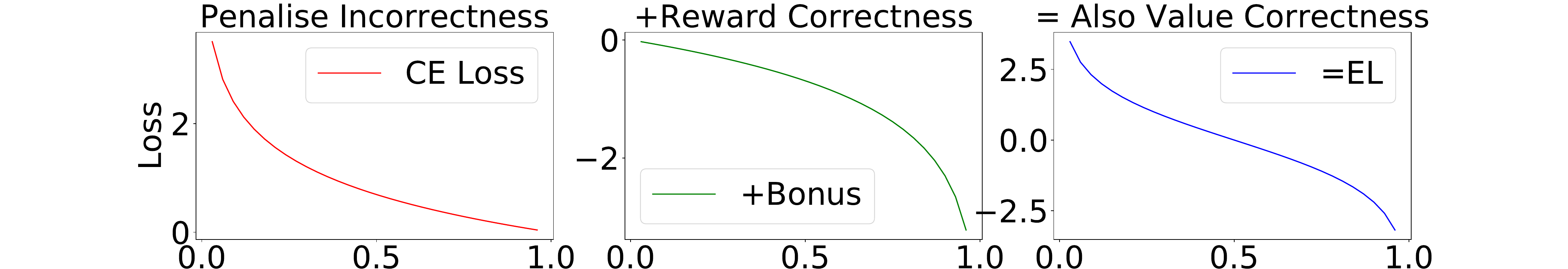}}\caption{Illustration of the encouraging loss, which is a counterexample. CE loss: $-logp$; Bonus: $log(1-p)$; Encouraging loss (EL): CE loss+ bonus. Bonus strengthens the learning of well-classified examples.}
\label{fig:ce_bonus_el}
\end{center}
\end{figure}

As plotted in the Figure \ref{fig:ce_bonus_el}, encouraging loss is the addition of CE loss and an additional loss (we term it as a \textbf{bonus}) that makes the loss steeper again when $p$ goes higher. 
The \textbf{normal bonus} is the mirror flip of the CE loss: $bonus=log(1-p)$, we clamp the value in the logarithm by a small epsilon (e.g., 1e-5) to avoid numerical instability. The encouraging loss with normal bonus is:
\begin{equation}
    \mathcal{L}_{EL} = -\log p_{\vtheta}(\vx)[\evy]) + log (1-p_{\vtheta}(\vx)[\evy]).
\end{equation}
We name it \textbf{encouraging loss} since it encourages the model to give more accurate predictions by rewarding these near-correct predictions with a bonus. As long as the additional bonus is concave, its steepness for larger $p$ is larger, indicating that the EL with that bonus pays more attention to the well-classified examples than the CE loss.

\begin{figure}[t]
\begin{center}
\centerline{\includegraphics[width=1.2\linewidth]{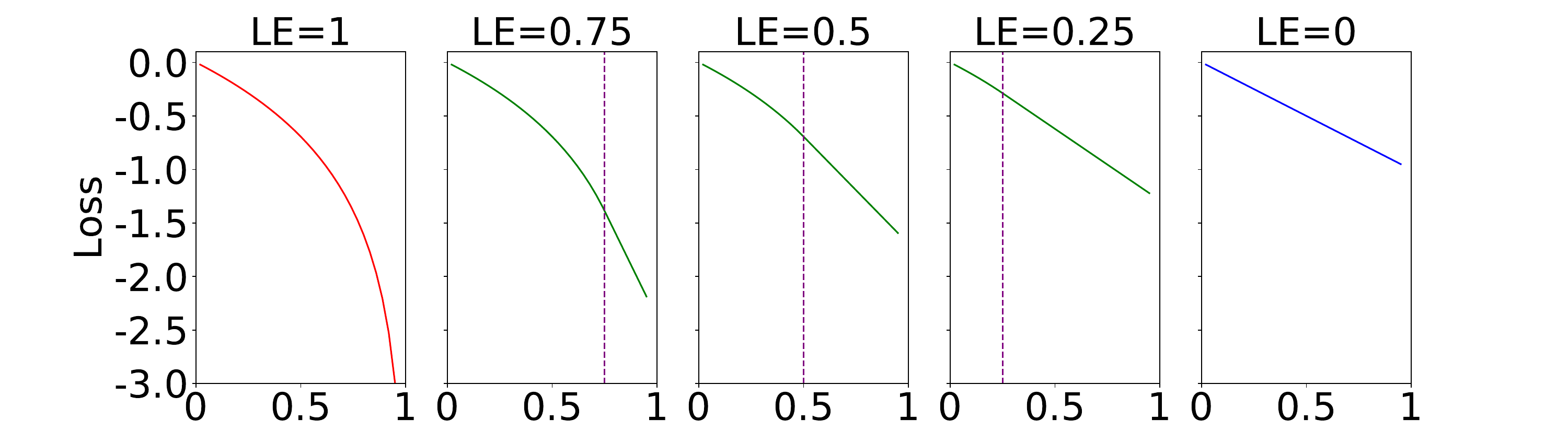}}
\caption{These are variations of bonuses for encouraging loss. (Left): normal bonus $log(1-p)$; (Others): conservative bonus. \textbf{$LE$} is the position where Log curve Ends in conservative log bonus, loss curve continues after $LE$ with the tangent at $p=LE$ of the log curve. These variations gradually increase steepness for well-classified examples from right to left. The original CE loss is the case $bonus=constant$.} 
\label{fig:vbonus}
\end{center}
\end{figure}

To make the gradient of EL be closer to CE to fit the existing optimization system, and study the relative importance of the learning for well-classified examples compared to other examples, we can adjust the relative steepness of the additional bonus. 
We design many types of \textbf{conservative bonus} that approximate the normal bonus but are more conservative and show them in Figure \ref{fig:vbonus}. These variants' \textbf{L}og curve \textbf{E}nds (\textbf{LE}) early in the high-likelihood region and replace the log curve with the tangent of the endpoint.  The relative importance of optimization for well-classified examples in encouraging loss with these bonuses is larger than CE and gradually increases from the right to the left. Let $p$ denote $p_{\vtheta}(\vx)[\evy])$,
\begin{equation}
\mathcal{L}_{EL_{LE}}  =\left\{
\begin{aligned}
&-\log p+ log (1-p) \quad p \leq LE.
\\
&-\log p - \frac{p-LE}{1-LE} +log(1-LE) \quad p > LE.\\
\end{aligned}
\right.
\end{equation}

The bonus can also be designed to be more aggressive than the normal bonus.


\subsection{The Counterexample Can Solve the Identified Issues in CE Loss} \label{sec: property}
Taking the normal bonus as a typical example, we analyze the benefits of learning with encouraging loss.

\paragraph{Encouraging loss enhances the representation learning from the part of well-classified examples} The steepness of the encouraging loss is $-\frac{1}{p}-\frac{1}{1-p}$, so the gradients are:
\begin{equation} \label{cf:el-gradient}
\begin{aligned}
\frac{\partial \mathcal{L}_{EL}}{\partial \vtheta} 
= -1\cdot \frac{\partial f_{\vtheta}(\vx)[y]}{\partial  \vtheta}.
\end{aligned}
\end{equation}
In contrast to gradients of NLL/CE loss in Eq. (\ref{cf:nll-gradient}), here gradients are independent of the likelihood $p$ (this is a bit like ReLU in DNNs~\cite{relu}). As to the EL with conservative bonus, since the bonus is concave, gradients are also larger for well-classified examples. For these conservative variants, gradients for probability before LE ($<$1) are the same as that when LE=1, and the gradients after LE ($<$1) are $\left(-\frac{1}{p} -\frac{1}{1-LE}\right)*\left[p\left(1-p\right)\right]$ times smaller, but they do not scale linearly as CE loss does.

\paragraph{Encouraging loss makes the model have smaller obstacles to reduce the energy on the data}
The conditional energy form of the EL is:
\begin{equation}\label{cf:el-energy}
    \begin{aligned}
    \mathcal{L}_{EL} = E_{\theta}(y\mid\vx) + log[ \sum_{y' \ne y} \exp(-E_{\vtheta}(y'\mid\vx))].
    \end{aligned}
\end{equation}
The difference between it and the conditional energy form of the CE loss in Eq. (\ref{cf:energy-ce}) lies in the second term. Notice that training with EL does not need to push up the energy on the data to minimize the second term, so there is more incentive to lower the energy on the data. Although conservative bonus $<log1$ and $>log(1-p)$ do not remove the obstacle in the second term, the obstacle will be smaller than CE loss.

\paragraph{Encouraging loss makes margins grow faster} The margin perspective of gradients w.r.t EL is:
\begin{equation}\label{cf:el-gradient-margin}
\begin{aligned}
\frac{\partial \mathcal{L}_{EL}}{\partial \vtheta} =\frac{\sum_{y' \ne y} A (\frac{\partial ( f_{\vtheta}(\vx)[y']- f_{\vtheta}(\vx)[y])}{\partial  \vtheta})}{\sum_{y' \ne y} A}.
\end{aligned}
\end{equation}
Now the growth speed for the margin is $\frac{1+\sum_{y' \ne y} A}{\sum_{y' \ne y} A}$ times faster than that during the training with the CE loss. When the model is getting better, the exponent of negative gap $A$ gets close to 0, the ratio becomes large and helps further increase the margin. The EL with a conservative bonus has a smaller ratio, but the ratio is still large than $1$, so the model still has incentive to increase the margin as it becomes larger.

\section{Practical Effect of Encouraging the Learning of Well-classified Examples}\label{sec:practice}
This section analyzes the practical effect of encouraging learning well-classified examples by applying the counterexample to various classification tasks and settings. 

\subsection{Experiment Setup}\label{sec:exp-setup}
In here we briefly clarify the experiment setup, please refer to the Appendix\footnote{Please refer to \url{https://arxiv.org/abs/2110.06537} for Appendix.} and the code for more details. For reliability, each result is the mean result of $5$ different runs with error bars. Especially, on graph datasets, each run contains $50$ different train, valid, test splits of the data (proportion is 0.8, 0.1, 0.1, respectively) since a recent study indicates that different dataset splits largely affect the test performance~\cite{graph-pitfalls}. For other tasks, we use their official data splits.

\paragraph{Image Recognition} It is a typical application of multi-class classification. In these tasks, we need to predict the category each image belongs to. We adopt four tasks MNIST, CIFAR-10, CIFAR-100, and ImageNet~\cite{imagenet}, the descriptions of the dataset are in Appendix.
For training, we borrow code from repositories with good reproduced accuracy and keep all their default settings unchanged.  Specifically, we train the CNN model from \citet{l-softmax} on MNIST, train ResNet-50~\cite{resnet} and EfficientNet-B0~\cite{tan2019efficientnet} on CIFAR-10 and CIFAR-100 using the code by Narumiruna
, train the ResNet-50 on ImageNet with the example code from PyTorch,
train the EfficientNet-B0 on ImageNet using the code from timm~\cite{rw2019timm}
. We choose ResNet-50 and EfficientNet-B0 because they are canonical and SoTA parameter-efficient models, respectively. For evaluation, we report the best top-1 accuracy on the test set following the common practice. 



\paragraph{Graph classification}
Typical applications of graph classification are to binary classify the functionality of the graph-structured biological data. We do the experiments on PROTEINS ($1113$ graphs of protein structures) ~\cite{dobson2003distinguishing,borgwardt2005protein} and  NCI1 ($4110$ graphs of chemical compounds) ~\cite{wale2008comparison}.
The model is the node feature learning model GCN~\cite{kipf2016semi} with the pooling method SAGPooling~\cite{lee2019self}. We report test accuracy on the early stopping model with the best valid accuracy.

\paragraph{Machine translation} In this task, we need to sequentially select a word class from the vocabulary, containing  tens of thousands of word classes. We perform experiments on IWSLT De-En ($160K$ training sentence pairs) and IWSLT Fr-En ($233K$ training sentence pairs). The base model is Transformer \cite{transformer}, and the evaluation metric is BLEU which calculates how many $N$-grams ($N$-gram is a contiguous sequence of $N$ classification predictions) both exist in the predicted sequence and the generated sequence~\cite{bleu}. We adopt most settings from fairseq, including training and evaluation. The only 
modification is that we tune the best hyper-parameters for the default loss (CE loss with label smoothing) and then use them for training models with encouraging loss for fair comparisons.

\paragraph{Imbalanced classification} We perform experiments on a large-scale natural imbalanced classification dataset--iNaturalist 2018, which has $8,142$ classes, $438K$ training samples, and $24K$ valid samples. The setting for training and evaluation is the same as \citet{Kang2020Decoupling}, including separately evaluating results on subsets of ``many-shot'', ``medium-shot'', and ``few-shot''.

\begin{figure*}
    \centering
\centering
\centerline{\includegraphics[width=0.92\linewidth]{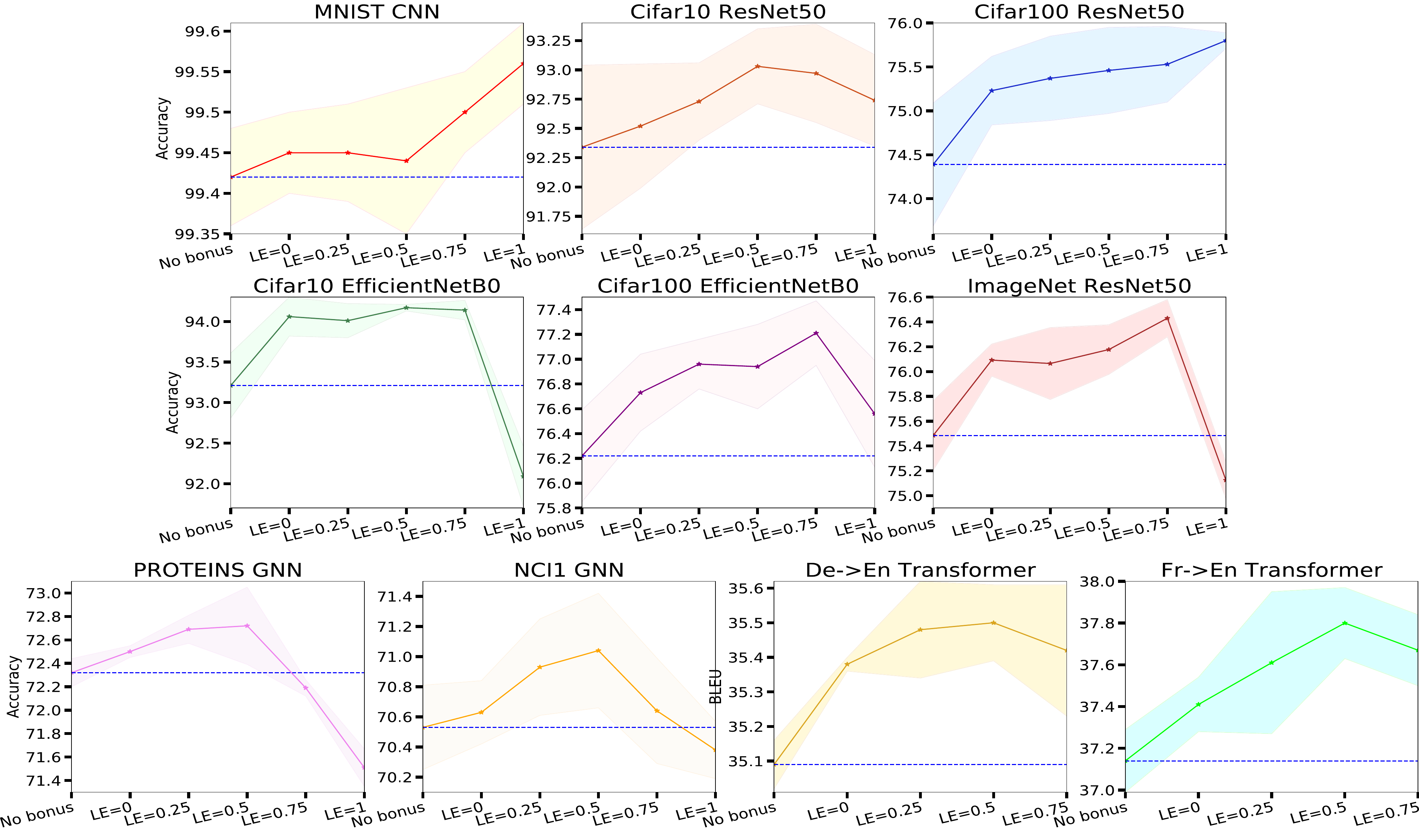}}
\caption{These figures plot performance under various settings. The colored areas denote the standard deviation of the 5 runs. We can see that enhancing the learning of well-classified examples can improve performance.} 
\label{fig:plot-all}
\end{figure*}

\begin{table}[]
    \centering
    \small
\begin{tabular}{@{}l|lllllllllll@{}} \toprule
Setting  & MNIST      & C10-r50    & C10-eb0           \\ \midrule
CE & 99.42$\pm$0.06 & 92.34$\pm$0.70 & 93.21$\pm$0.40   \\
EL       & 99.56$\pm$0.05 & 92.97$\pm$0.42 & 94.24$\pm$0.17  \\ \midrule

Setting & C100-r50   & C100-eb0   & Img-r50 \\ \midrule
CE& 74.39$\pm$0.70 & 76.22$\pm$0.37 & 75.49$\pm$0.28 \\
EL & 75.80$\pm$0.09 & 77.21$\pm$0.26 & 76.43$\pm$0.15 \\ \midrule
Setting    & Img-eb0    & Proteins   & NCI1       \\ \midrule
CE & 77.80$\pm$0.15 & 72.32$\pm$0.12 & 70.53$\pm$0.28  \\
EL & 78.28$\pm$0.13 & 72.76$\pm$0.18 & 71.04$\pm$0.38 \\ \midrule
Setting & De-En      & Fr-En \\ \midrule
CE& 35.09$\pm$0.07 & 37.10$\pm$0.06\\
EL& 35.50$\pm$0.11 & 37.73$\pm$0.17 \\
\bottomrule
\end{tabular}
    \caption{We summarize the results of CE and EL on various tasks. Encouraging the Learning of well-classified examples brings consistent improvements. We abbreviated the names, e.g., Img refers to ImageNet.}
    \label{tab:res-number}
\end{table}




\subsection{Improving Classification Performance}\label{sec:acc}

\paragraph{Image recognition of multi-class classification on the pixel data} 
We plot the results in Figure \ref{fig:plot-all} and summarize the improvements in Table~\ref{tab:res-number}. For EfficientNet-B0 on Imagenet, we directly run the baseline and encouraging loss with a conservative bonus (LE=0.75) to save energy since EfficientNet is not time-efficient. From all results, we point out that strengthening the learning of well-classified examples with an additional bonus can help improve the accuracy. 

Take ImageNet as an example, by strengthening the learning of well-classified examples, the accuracy of the canonical model ResNet-50 is improved by $0.94$ points, and the accuracy of the SoTA parameter efficient model EfficientNet-B0 can also be improved from $77.8$ to $78.28$. We want to point out that the improvement by EL is actually remarkable compared with other methods. For example: On ImageNet (1.28M images), we
improved the valid accuracy of EfficientNet-b0 from 77.8
to 78.3. Noisy Student~\cite{noisy-student} is the SoTA training method for
EfficientNet-b0, which enhances the performance from 77.3
to 78.1 while it relies on training on a much larger dataset,
google’s JFT-300M. Mixup~\cite{zhang2018mixup} improves the ResNet50 by 0.2
with the same number of epochs (90 epochs) and improves
by 1.5 with more than 2$\times$ epochs, while we can improve the
ResNet-50 by 0.9 without training extra epochs.


\paragraph{Binary graph classification on graph-structured data}\label{sec:graph} We can see from the Figure \ref{fig:plot-all} that strengthening the learning of well-classified examples with a conservative bonus can bring an increase in the accuracy of $0.44$ on Proteins and accuracy of $0.51$ on NCI1.



\paragraph{Machine translation of sequential multi-class classification on the text data} \label{sec:translation}
As we can see from the last two sub-figures of Figure~\ref{fig:plot-all}, on machine translation, rewarding correct predictions can bring BLEU score improvement of $0.41$ on De-En translation and $0.63$ on Fr-En translation. 
We show in the Appendix that our improvements are additive label smoothing~\cite{label-smoothing} and on par with it.

\begin{table}[]
\small
\centering
\begin{tabular}{l|l|l}
\toprule
Setting      & Cifar10-resnet50    &Cifar100-resnet50  \\ \midrule
CE     & 92.34$\pm$0.70 & 74.39$\pm$0.70 \\ 
CE (2xLR) &91.53$\pm$0.19 & 73.57$\pm$0.66 \\ \midrule
EL           & 92.74$\pm$0.38 & 75.80$\pm$0.09 \\
EL (0.5xLR) & 93.69$\pm$0.25 & 76.37$\pm$0.29 \\ \bottomrule
\end{tabular}
\caption{Training with EL (LE=1) can benefit from reducing the amount of gradients since the overall gradients of EL are larger than CE in theory.}
\label{tab:lr}
\end{table}
\paragraph{Discussion about the hyper-parameter LE and the conservative bonus}
Results from Figure \ref{fig:plot-all}  indicate that in many conservative settings (LE$\leq$0.5) where encouraging loss already pays more attention to well-classified examples than CE loss,  EL with conservative bonuses consistently improves the performance. However, encouraging loss with a steeper bonus can not further improve the accuracy of deep classification models in some settings. We compare EL with conservative bonus (LE$<$1) and normal bonus (LE=1) below.

\emph{a. Why does a steep bonus like LE=1 does not bring improvements in some scenarios?} The reason is that CE loss implicitly decreases the learning rate along with the training as the gradient norm of CE loss decays during training, and existing methods for optimization which adapt to CE loss should be modified. In our experiments, we choose to adopt the best setting in baselines for EL, which may not be the most suitable for encouraging loss. To verify this, we first show in Table~\ref{tab:lr} that when we enlarge the learning rate for CE loss, accuracy also drops. Then, we find that re-normalizing the gradients of encouraging loss by decreasing the global learning rate can help learn better from well-classified examples. For example, we can continue improving the accuracy gap between CE loss and encouraging loss to $1.35$ on Cifar10-resnet50 and $1.98$ on Cifar100-resnet50, respectively. 

\emph{b. Every performance peak seems to occur between LE=0.5 and LE=0.75.} 
These two settings strengthen the training for well-classified examples while not much changing the overall gradient norm. For example, in our preliminary experiments, we observe that on Transformer translation, LE=1 makes the norm of gradients $>5\times$ larger and brings 1-2 BLEU drop, but the norm is only $<1.7\times$ larger than CE for LE=0.75. Thus, when we directly use the original training hyper-parameters for CE loss for a fair comparison, LE=0.5 and LE=0.75 are good default choices. 

\emph{c. How to select the additional hyper-parameter LE?} First, practitioners can choose LE=$0.5$ as it works consistently better than CE loss on all tasks with existing mainstream systems we tried. They can select higher LE for models that are stable to train (e.g., ResNet). Second, Table~\ref{tab:lr} shows that we can benefit from modifying the original system to use higher LE (e.g., LE=$1$), which yields better results.

\paragraph{Easy to apply}
Besides all the promising results shown above, our method has another advantage in that it can be widely applicable as a direct plug-in module without changing all the original hyper-parameters.

%

\subsection{Addressing the Issues of CE Loss in Practice} \label{sec:margin}
In this subsection, we demonstrate that strengthening the learning of well-classified examples addresses the issues that CE loss faces as we discussed in \S \ref{subsec: limitation of ce}. Due to the space limit, we compare results between CE loss and EL with a normal bonus for training ResNet-50 on CIFAR-10. Please refer to the Appendix for more results in other settings.

\paragraph{Representation Learning}
Besides the overall accuracy improvement illustrated in \S \ref{sec:acc}, we also perform experiments to evaluate the learned representations by two losses. Specifically, we first use CE loss and EL to train the whole model, and then only train the re-initialized output layer with CE loss but fix the representations. The procedure is similar to \citet{Kang2020Decoupling} that decouples the training into the representation learning phase and classification phase. We observe that representation learning with EL achieves an accuracy of \textbf{$92.98\pm0.01$},  but the representation learning with CE loss only gets an accuracy of \textbf{$91.69\pm0.04$}.

\begin{figure}
    \centering
\begin{center}
\centerline{\includegraphics[width=1.0\linewidth]{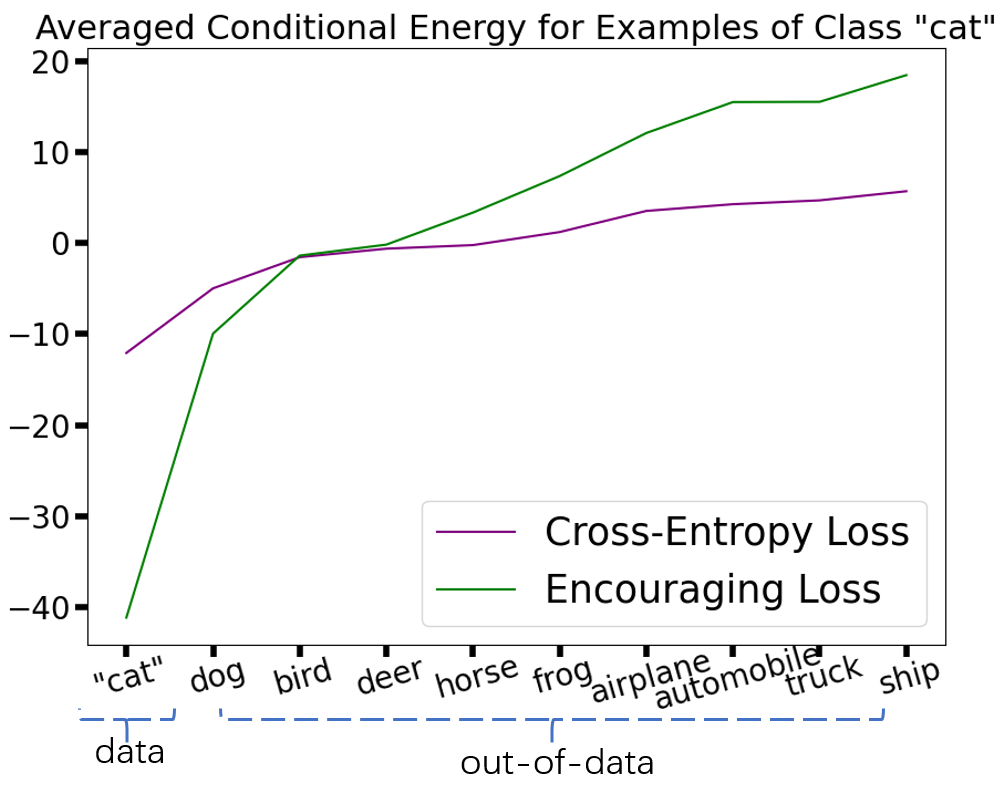}}
\caption{Conditional energy $E(y\mid x)$ averaged over the examples($X_{y*}$) of target class (here it is 'cat'), the energy around the data ($y=y*$) becomes sharper with EL.} 
\label{fig:energy}
\end{center}
\end{figure}

\paragraph{Energy optimization} We plot the conditional energy $E(y \mid x)$ averaged over the examples of class ``cat'' in Figure \ref{fig:energy}. We can see that the energy around the data becomes sharper with the help of encouraging loss since it pushes down the energy on the data more than CE loss.

\begin{figure}
    \centering
\centerline{\includegraphics[width=1.0\linewidth]{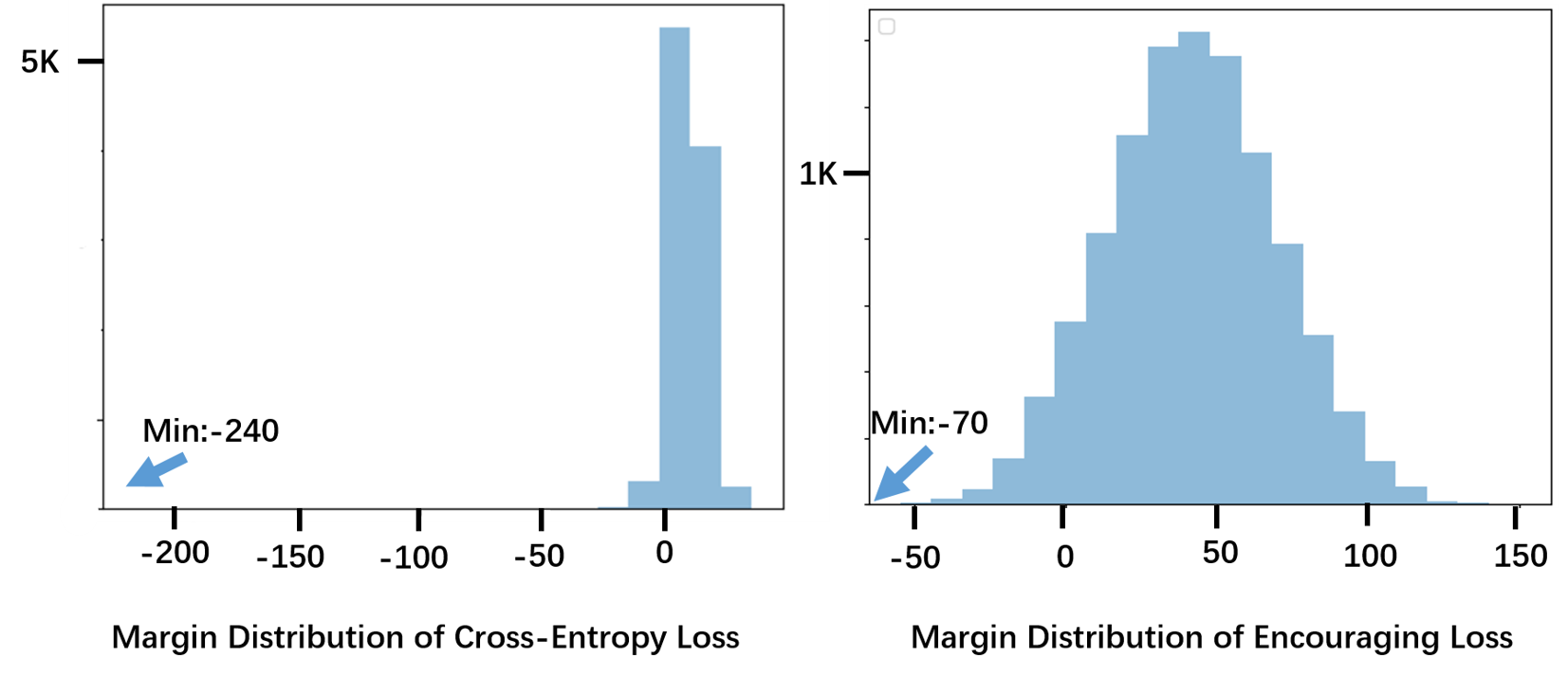}}
\caption{(Left): Margins of CE loss distributed among -240 to 50, centered by 15. (Right): Margins of EL distributed among -70 to 170, centered by 50. The learning of well-classified examples help double the margins in classification. } 
    \label{fig:margin}
\end{figure}

\paragraph{Growth of margins} We can see from Figure \ref{fig:margin} that margins of encouraging loss are several times larger than margins of CE loss. These results demonstrate that learning well-classified examples with the additional bonus greatly improves the classification models by enlarging margins, which further makes the model has great generalization and robustness as we will discuss in the following.

\subsection{Coping with Complex Application Scenarios} \label{sec:complex}
This section shows that enhancing the learning of well-classified examples by EL cope well with three complex application cases since this idea mitigates the three issues.

\paragraph{Imbalanced classification} \label{sec:imb}
In imbalanced classification, we improve the classification performance of rare classes with the help of representation learning from samples of data-rich classes~\cite{Kang2020Decoupling}. Because enhancing the learning of easy samples also enhances their representation learning, we believe this property benefits imbalanced learning. 
 We conduct validation experiments on the iNaturalist 2018 dataset, and the results are in Table~\ref{tab:imb}. We find that encouraging the learning of well-classified samples makes models' performance outperform that trained with CE loss, both for the case with conservative bonus or the case with aggressive bonus (only to reward the highly well-classified samples of $p>0.5$ with the normal bonus). We can also combine our idea with other advanced methods~\cite{LDAM, Kang2020Decoupling}. For the additive experiment with ``Decoupling classifier and representation learning with CRT'' \cite{Kang2020Decoupling}, in the representation learning phase, we not only remain the learning of data-rich classes as they do but also revive the learning of well-classified examples. In the classifier learning phase, we keep all the settings unchanged. Our method improves them by 1.9 points, \textbf{which empirically verifies that the traditional re-weighting at the sample level (CE loss down-weights the importance of well-classified samples) is also harmful to the representation learning}.
 
\begin{table}[t]

\centering
\makeatletter\def\@captype{table}\makeatother 

\begin{tabular}{lllll}
\toprule
\small
\multirow{1}{*}{Method}  & \multicolumn{4}{c}{iNaturalList2018}                  \\
                                              & ALL & Many & Med. & Few \\ \midrule
CE loss & 64.3 &74.1 &65.9 &59.8               \\ 
+ Conservative Bonus &65.3 &74.3 &66.6 &61.2 \\ 
+ Normal Bonus  & 65.8 &74.4 &66.6 &62.4 \\
+ Aggressive Bonus & 66.3 & \textbf{75.1}    & 67.4      & 62.6   \\ \midrule
Decoupling Reps\&Cls\ &64.9 &71.4 &65.9 &61.9  \\
+ Normal Bonus &66.8 & 71.7 & 67.6 & 64.6  \\ 
\midrule
Deferred Re-weighting &68.1 &71.0 & 68.3 & 67.1 \\
+ Normal Bonus &\textbf{70.3} &69.0& \textbf{70.1} &	\textbf{70.9} \\ 
\bottomrule
\end{tabular}

\caption{Comparison between CE loss and encouraging loss (add a bonus to CE loss) on imbalanced classification dataset iNaturalist 2018, the average  standard deviation for results on iNaturalList 2018 is 0.4. Additional bonuses that revive the learning of well-classified examples bring improvements both on CE loss and its advanced variants. LE of the conservative bonus here is 0.5. Med is Medium.}
\label{tab:imb}

\end{table}

\begin{figure}[t]
    \centering
    \includegraphics[width=0.72\linewidth]{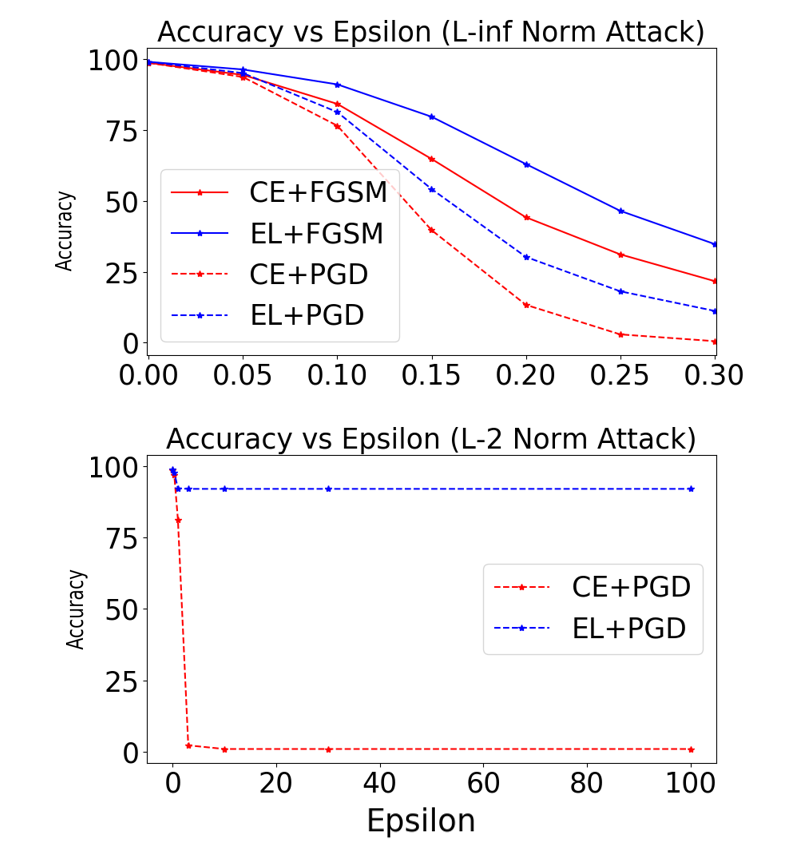}
    \caption{Accuracy of CE loss and encouraging loss under adversarial attacks of FGSM and PGD. }
   \label{fig:attack}
\end{figure}

\paragraph{OOD detection}
The task of OOD detection is to detect whether a new sample belongs to the distribution of training data. 
We can improve the performance of OOD detection since EL sharpens the energy surface around the data.
For OOD detection, we use an indicator to detect whether samples are in the distribution (In-D) of the training set or out of the distribution (OOD). We use the minimum conditional energy $min_{y' \in\mathbb{Y}} E_{\vtheta}(y' \mid \vx)$ as the indicator.
We show in the Appendix that our indicator is better than the maximum probability and free energy~\cite{free-energy}. 
We conduct an experiment on a small-scale OOD detection task (In-D: MNIST, OOD: Fashion-MNIST) and a large-scale task (In-D: ImageNet, OOD: iNautralist2018). Samples in Fashion-MNIST are clothes, while samples in MNIST are digital numbers; samples in ImageNet are instances of objects, while samples in iNautralist2018 are instances of species (refer to Appendix for detailed settings).
 We can see from Table~\ref{tab:ood} that EL leads to better performance than CE loss. For example, we reduce the metric FPR95 from $8.10\%$ to $0.95\%$ on MNIST vs. Fashion MNIST. These results confirm that strengthening the learning of well-classified examples leads to discriminative energy distribution.
\begin{table}[]
\small
\begin{tabular}{l|ll|ll} \toprule
  Setting & \multicolumn{2}{l}{MNIST vs. F. MNIST} & \multicolumn{2}{|l}{Img vs .iNautralist2018} \\ 
Metric& AUROC$\uparrow$                 & FPR95$\downarrow$                 & AUROC$\uparrow$                   & FPR95$\downarrow$                   \\ \midrule
CE & 95.68                 & 20.92                 & 76.54                   & 75.40                    \\
EL & 98.04                 & 8.69                  & 78.41                   & 71.86 \\ \bottomrule                  
\end{tabular}
\caption{EL significantly improves OOD detection performance. We use F. MNIST to denote Fashion MNIST. Higher AUROC and lower FPR95 are better.}
\label{tab:ood}
\end{table}

\paragraph{Robustness to adversarial attacks}
We have shown in theory and practice that paying more attention to the classified samples can significantly increase margins. This property is likely to increase the robustness of adversarial examples because the predictions of disturbed samples are likely to remain far away from the decision boundary. To verify this assumption, on MNIST, we use the FGSM~\cite{fgsm},  PGD~\cite{pgd} with $L_\infty$ bound, and PGD with $L_2$ bound to construct adversarial samples to attack the models trained with CE loss and with EL, respectively. The results in Figure \ref{fig:attack} validate our assumption that the model trained with EL is more robust on predictions when receiving a specific range of adversarial disturbances.

\subsection{Well-classified Examples Are also Underestimated in MSE}
\begin{table}[t]

  \centering
  \small
  \begin{tabular}{lll}
    \toprule
    \cmidrule(r){2-3}
    Name     & Origin Loss     & +Mirror Bonus \\
    \midrule
    CE loss + softmax &99.42 $\pm$ 0.06 &99.56 $\pm$ 0.05 \\
    MSE + softmax & 99.56 $\pm$ 0.05  & 99.66 $\pm$ 0.04     \\
    MSE + sigmoid & 99.63 $\pm$ 0.04 & 99.69 $\pm$ 0.02      \\
    \bottomrule
  \end{tabular}
\label{tab:mse}
  \caption{Well-classified examples (the learning  of them is encouraged in the Mirror Bonus) also help improve MSE, which is better than CE loss on MNIST.}
    \label{tab:mnist-mse}
\end{table}
It is well-known to the community that CE loss outperforms MSE for training deep neural networks, and the superiority has also been proven to come from the steeper gradient of CE~\cite{back:ce-steep}. However, Table~\ref{tab:mnist-mse} shows that MSE beat CE loss on MNIST. The possible reason is that CE loss suffers from a much higher sensitivity to bad-classified examples.  Nevertheless, we find that by adding a mirror bonus $-(y*p)^2-((1-y)*(1-p))^2$ to MSE $(y-p)^2$ to encourage the learning of well-classified examples, the performance improves. This indicates that MSE also has the problem of underestimating well-classified examples.


\section{Limitations} \label{sec: limitations} %
To facilitate future research, we analyze the difficulties and the possible solutions in this new area.
First, EL with the normal bonus can not improve the accuracy of deep classification models in some settings. One possible reason is that  CE loss reduces the overall gradient norm, and existing optimization which adapts to CE loss should be modified. We show in Table~\ref{tab:lr}  that reducing the learning rate when employing encouraging loss is a possible solution.
Second, models trained with EL tend to give more confident predictions, and the Expected Calibration Error (ECE) ~\cite{ECE} is slightly increased. Nevertheless, we can mitigate the issue by combining EL with label smoothing~\cite{label-smoothing} to ECE. We show in the Appendix that the combination of EL and label smoothing gets lower ECE than CE or CE with label smoothing.


\section{Conclusion}
In this paper, we theoretically and empirically show that well-classified examples are helpful for the further improvement of deep classification models. To illustrate this finding, we first directly analyze the failure of common practice, which weakens the learning of these examples. We then propose some counterexamples that value the learning of well-classified examples, and verify their positive effects in optimization through various and comprehensive experiments.

\section*{Ethical Statement}
In this work, we do not introduce new datasets but use the existing widely-used datasets from their public release. As to societal harm, we do not facilitate social bias, but we can improve the performance of minority classes. Since algorithms based on our idea have a similar calculation time with CE loss, we do not facilitate cost more energy. Thus, our work does not have potential ethical considerations.

\section*{Acknowledgments}
This work is partly supported by National Key R\&D Program of China (2020AAA0105200), Natural Science Foundation of China (NSFC) No. 62176002, and Beijing Academy
of Artificial Intelligence (BAAI).  We thank all the
anonymous reviewers for their constructive comments and
Zhiyuan Zhang, and Sishuo Chen for helpful discussion in preparing the manuscript.

\bibliography{aaai22}

\appendix
\section{Complete Calculation of Equations}
Recall that the NLL loss is:
\begin{equation*}\label{cf:nll-ap}
\begin{aligned}
\mathcal{L}_{NLL}
&= -\log p_{\theta}(\evy\mid\vx) = - \log p_{\theta}(\vx)[\evy]
\end{aligned}
\end{equation*}
In the following paragraphs, we simply use $p$ to denote the likelihood of the target $p_{\theta}(\vx)[\evy]$.

In Eq. \ref{cf:nll-gradient}, we calculate gradients for the model parameters with the CE loss, its complete calculation is:
\begin{equation*} \label{cf:nll-gradient-ap}
\begin{aligned}
\frac{\partial \mathcal{L}_{NLL}}{\partial \vtheta} 
&=  \frac{\partial \mathcal{L}_{NLL}}{\partial p}  \frac{\partial \sigma ( f_{\vtheta}(\vx)[y])}{\partial f_{\vtheta}(\vx)[y] } \frac{\partial f_{\vtheta}(\vx)[y]}{\partial  \vtheta} \\&= -\frac{1}{p} [p(1-p)] \frac{\partial f_{\vtheta}(\vx)[y]}{\partial  \vtheta}=(p-1) \frac{\partial f_{\vtheta}(\vx)[y]}{\partial  \vtheta}.
\end{aligned}
\end{equation*}

In Eq. \ref{cf:energy-ce} we re-write the CE loss, its complete calculation is:
\begin{equation*}\label{cf:energy-ce-ap}
\begin{aligned}
    \mathcal{L}_{NLL} &= -f_{\vtheta}(\vx)[y] \\&+\log [\exp(f_{\vtheta}(\vx)[y]) + \sum_{y' \ne y} \exp(f_{\vtheta}(\vx)[y'])] \\
    &= E_{\theta}(y \mid \vx) + \log[\exp(-E_{\vtheta}(y\mid\vx)) \\&+ \sum_{y' \ne y} \exp(-E_{\vtheta}(y' \mid \vx))] ,
\end{aligned}
\end{equation*}
in which the we regard the negative of the logit as the conditional energy: $E_{\theta}(y \mid \vx)=-f_{\vtheta}(\vx)[y]$.

Note that from the view of margin, the NLL loss can also be written as $\mathcal{L}_{NLL} =\log [\frac{\exp(f_{\vtheta}(\vx)[y])}{\exp(f_{\vtheta}(\vx)[y])}  + \frac{\sum_{y' \ne y} \exp(f_{\vtheta}(\vx)[y'])}{\exp(f_{\vtheta}(\vx)[y])}] =\log[1+\sum_{y' \ne y} \exp(f_{\vtheta}(\vx)[y'] -f_{\vtheta}(\vx)[y])]$.

In Eq. \ref{cf:ce-gradient-margin}, we use $A$ to denote $\exp(f_{\vtheta}(\vx)[y'] -f_{\vtheta}(\vx)[y])$, and calculate the gradients of the NLL loss w.r.t. the parameter $\vtheta$:
\begin{equation*}\label{cf:ce-gradient-margin-ap}
\begin{aligned}
\frac{\partial \mathcal{L}_{NLL}}{\partial \vtheta} =\frac{\sum_{y' \ne y} A (\frac{\partial ( f_{\vtheta}(\vx)[y']- f_{\vtheta}(\vx)[y])}{\partial  \vtheta})}{1+\sum_{y' \ne y} A}
\end{aligned}
\end{equation*}

The normal bonus is the mirror flip of the CE loss, it is $bonus=\log(1-p)$, and the Encouraging Loss (EL) with normal bonus is:
\begin{equation*}
    \mathcal{L}_{EL} = -\log p_{\vtheta}(\vx)[\evy]) + \log (1-p_{\vtheta}(\vx)[\evy]))
\end{equation*}

In Eq. \ref{cf:el-gradient} we calculate the gradients of the encouraging loss with normal bonus, since the steepness $\nicefrac{\partial \mathcal{L}}{\partial p}$ of this loss is $-\frac{1}{p}-\frac{1}{1-p}$, its complete calculation is:
\begin{equation*} \label{cf:el-gradient-ap}
\begin{aligned}
\frac{\partial \mathcal{L}_{EL}}{\partial \vtheta} &=  \frac{\partial \mathcal{L}_{EL}}{\partial p}  \frac{\partial \sigma ( f_{\vtheta}(\vx)[y])}{\partial f_{\vtheta}(\vx)[y] } \frac{\partial f_{\vtheta}(\vx)[y]}{\partial  \vtheta} \\&= (-\frac{1}{p}-\frac{1}{1-p}) [p(1-p)] \frac{\partial f_{\vtheta}(\vx)[y]}{\partial  \vtheta} \\
&= -(1-p +p)\cdot \frac{\partial f_{\vtheta}(\vx)[y]}{\partial  \vtheta} 
= -1\cdot \frac{\partial f_{\vtheta}(\vx)[y]}{\partial  \vtheta}
\end{aligned}
\end{equation*}
As to the EL with conservative bonus, since the bonus is concave, gradients are also larger for well-classified examples.

In Eq. \ref{cf:el-energy}, we write the  conditional energy form of the EL, its complete calculation is:
\begin{equation*}\label{cf:el-energy-ap}
    \begin{aligned}
    \mathcal{L}_{EL} &=-\log p_{\vtheta}(\vx)[\evy]) + \log (1-p_{\vtheta}(\vx)[\evy])) \\
    &= -f_{\vtheta}(\vx)[y] + log[ \sum_{y' \in \mathbb{Y}} \exp(f_{\vtheta}(\vx)[y'])]  \\&+ \log[\sum_{y' \ne y} \exp(f_{\vtheta}(\vx)[y'])] - \log[\sum_{y' \in \mathbb{Y}} \exp(f_{\vtheta}(\vx)[y'])] \\
    &= -f_{\vtheta}(\vx)[y] + \log[ \sum_{y' \ne y} \exp(f_{\vtheta}(\vx)[y'])]  \\&= E_{\theta}(y\mid\vx) + \log[ \sum_{y' \ne y} \exp(-E_{\vtheta}(y'\mid\vx))]
    \end{aligned}
\end{equation*}
    The difference between Eq. (8) and the conditional energy form of the CE loss in Eq. (4) lies in the second term, notice now we do not need to push up energy on the data to minimize the second term, so there is more incentive to lower the energy on the data. Since the conservative bonus is  $<log1=log[\sum_{y' \in \mathbb{Y}} \exp(f_{\vtheta}(\vx)[y'])] - log[\sum_{y' \in \mathbb{Y}} \exp(f_{\vtheta}(\vx)[y'])]$ and $>log(1-p)=log[\sum_{y' \ne y} \exp(f_{\vtheta}(\vx)[y'])] - log[\sum_{y' \in \mathbb{Y}} \exp(f_{\vtheta}(\vx)[y'])]$, although it does not remove the obstacle in the second term, the obstacle is smaller than CE loss.
    
Note that from the view of margin, the EL loss with normal bonus can also be written as $\mathcal{L}_{EL} = \frac{\sum_{y' \ne y} \exp(f_{\vtheta}(\vx)[y'])}{\exp(f_{\vtheta}(\vx)[y])}] =log[\sum_{y' \ne y} \exp(f_{\vtheta}(\vx)[y'] -f_{\vtheta}(\vx)[y])]$.

In Eq. \ref{cf:el-gradient-margin}, we calculate the gradients w.r.t EL from the view of margin:
\begin{equation*}\label{cf:el-gradient-margin-ap}
\begin{aligned}
\frac{\partial \mathcal{L}_{EL}}{\partial \vtheta} =\frac{\sum_{y' \ne y} A (\frac{\partial ( f_{\vtheta}(\vx)[y']- f_{\vtheta}(\vx)[y])}{\partial  \vtheta})}{\sum_{y' \ne y} A}
\end{aligned}
\end{equation*}
Growth speed for the margin is ratio $\frac{1+\sum_{y' \ne y} A}{\sum_{y' \ne y} A}$ times faster than that during the training with the CE loss. When the model is getting better, the exponent of negative gap $A=\exp(f_{\vtheta}(\vx)[y'] -f_{\vtheta}(\vx)[y])$ gets close to $0$, the ratio becomes large and helps further increase the margin. The EL with a conservative bonus has a smaller ratio, but the ratio is larger than $1$ since the denominator of the ratio is larger.

\section{Details of the Experiment Setups} \label{sec:app-setting}
In this section, we describe the details of the experiments. Please refer to the code for reproduction. We provide the code in the supplementary material.

\subsection{Experiment Setups for Main Results}
In this subsection, we give details of experiment setups for main results and other experiments with the same settings.

For the results in image recognition, machine translation,imbalanced classification, each result is the mean result of $5$ different runs with error bars to ensure reliable results. For graph classification, each run contains $50$ different train, valid, test splits of the data (proportion is 0.8, 0.1, 0.1 respectively) since a recent study indicates that different dataset splits largely affect the test performance~\cite{graph-pitfalls}. For other tasks, we use their official data splits. 

We perform all experiments on the commonly used datasets and get data from their public release.  Except for the graph data set, which ended training through the early exit, our method's number of training rounds is the same as the baseline for other data sets.

\subsubsection{Image recognition} It is a typical application of multi-class classification. In these tasks, we need to predict the category an image belongs to. In this subsection, we mainly discuss tasks of balanced image recognition in which every class has almost the same number of samples.

For datasets, we adopt four tasks MNIST\footnote{MNIST data: \url{http://yann.lecun.com/exdb/mnist/}}, CIFAR-10\footnote{CIFAR-10 / 100 data: \url{https://www.cs.toronto.edu/~kriz/cifar.html}}(It's license is unknown in paperwithcode), CIFAR-100, and ImageNet~\cite{imagenet}\footnote{ImageNet data: \url{https://image-net.org/challenges/LSVRC/2012/}}, they have $10$, $10$, $100$ and $1000$ classes respectively and $60K$,$50K$,$50K$,$1.2M$ training samples respectively.

For models, we borrow the code from several repositories with good reproduced accuracy.  Specifically, we adopt the  CNN model from \citet{l-softmax} on MNIST, ResNet-50~\cite{resnet} and EfficientNet-B0~\cite{tan2019efficientnet} on CIFAR-10 and CIFAR-100 using the code by Narumiruna\footnote{Environments of experiments on CIFAR 10/100: \url{https://github.com/narumiruna/efficientnet-pytorch}}, we also ResNet-50 on ImageNet with the code from the official PyTorch examples for imagenet\footnote{Environments of experiments of ResNet-50 on ImageNet: \url{https://github.com/pytorch/examples/tree/master/imagenet}}, train the EfficientNet-B0 on ImageNet using the code from timm~\cite{rw2019timm} since this code-base publicly releases scripts that can replicate a good performance for Efficientnet-B0.  We choose ResNet-50 and EfficientNet-B0 because they are the common-used model and the SoTA parameter-efficient model, respectively. 

We keep all the default hyper-parameters unchanged and consistent with that in the source repositories for preprocessing, training, and evaluation.  Some of these hyper-parameters are: For training, we use RMSProp\footnote{RMSprop is an unpublished, adaptive learning rate method proposed by Geoffrey Hinton in Lecture 6e of his Coursera Class} to minimize the CE loss or the encouraging loss for EfficientNet-B0 on ImageNet since RMSProp is the default optimizer for training EfficientNet on ImageNet, and use SGD~\cite{sgd}  for other settings. The learning rate is $0.1$ with the batch size of $256$ where we follow the common practice, and readers can find the learning rate schedule in the code. 



\subsubsection{Graph Classification}
Typical applications of graph classification are to binary classify the functionality of the graph structured biological data,  we do the experiments on PROTEINS ($1113$ graphs of protein structures) ~\cite{dobson2003distinguishing,borgwardt2005protein} and  NCI1 ($4110$ graphs of chemical compounds) ~\cite{wale2008comparison}.
The model is the combination of the Graph Convolution (GCN)~\cite{kipf2017semisupervised} model for node feature learning and the pooling method SAGPooling~\cite{lee2019self} for selecting representative nodes in each graph. The details about the architecture can be found in the paper of SAGPooling~\cite{lee2019self} and our code. Our code and hyper-parameter settings are borrowed from pytorch-geometric~\cite{pg}, and we do not change the hyper-parameters there. Specifically, for training, we adopt the default optimizer Adam~\cite{adam} with a batch size of $128$ and a learning rate of $5\times 10^{-4}$. For evaluation, we report test accuracy on the early stopping model with the best valid accuracy. The patience is set to $50$ following SAGPooling~\cite{lee2019self}. The best epochs on the valid set are $35$ and $140$ respectively, so $85$ and $190$ epochs are executed. 

\subsubsection{Machine Translation} In this task, we need to sequentially select one word class from the vocabulary which contains tens of thousands of word classes at each position. We perform experiments on IWSLT De-En\footnote{De-En translation data: We get the data from \url{https://workshop2014.iwslt.org/}} ($160K$ training sentence pairs, the target language has a vocabulary of $6632$ words) and IWSLT Fr-En\footnote{Fr-En translation data: We get the data from \url{https://workshop2017.iwslt.org/}} ($233K$ training sentence pairs, the target language has a vocabulary of $16240$ words). The model is Transformer \cite{transformer}.  We adopt most settings from fairseq \cite{ott2019fairseq}, including preprocessing, training, and evaluation. The only difference is that we tune the best hyper-parameters for the default loss (It is the CE loss with label smoothing. ) and then use them to train with encouraging loss.  Specifically, we use Adam optimizer, the highest learning rate is $1e-3$, on De-En translation, we train the model for $20K$ updates with each batch has $16K$ tokens, on Fr-En translation, we train the model for $100K$ updates with each batch has $4K$ tokens. The number of updates is determined on the valid set, and the criteria is that the last $10$ epochs have similar valid performance.
For evaluation, the metric is BLEU which calculates how many $N$-grams ($N$-gram is a contiguous sequence of $N$ classification predictions) both exist in the predicted sequence and the generated sequence~\cite{bleu}, we use tokenized BLEU with the default tokenize method. Following common practice~\cite{transformer}, we report the BLEU of the averaged checkpoint over checkpoints saved in the last $10$ epochs.

\subsection{Experiment Setups for Verifying the Properties} 
In Sections 4.3, we empirically verify the theoretical results that well-classified examples help representation learning, energy optimization, and the growth of margins.  Since the last two are straightforward, we describe the details for the first verification.

\subsubsection{Representation Learning in Balanced Classification} In Section 4.3 of the main paper, we report the representation learning performance on balanced classification. Specifically, we first use CE loss and EL to train the whole model and then only train the re-initialized output layer with CE loss but fix the representations. The procedure is similar to \citet{Kang2020Decoupling} that decouples the training into the representation learning phase and classification phase.  The additional stage of retraining the classifier took $20$ epochs, starting from $0.1$ and decreasing by $0.1$ times every $5$ epochs. We selected these hyper-parameters because they are the most straightforward hyper-parameter setting from our perspectives. In our initial experiment, there were similar results in other settings.

\subsection{Experiment Setups for Complex Application Scenarios}
\subsubsection{Imbalanced Classification} We also show that well-classified examples help in representation learning on imbalanced classification.  We perform  experiments on a large-scale  natural imbalanced classification dataset--iNaturalist 2018\footnote{iNaturalist 2018 data: \url{https://github.com/visipedia/inat_comp/tree/master/2018}}, which has $8,142$ classes, $438K$ training samples and $24K$ valid samples. The setting for training and evaluation is the same as \citet{Kang2020Decoupling}, including using the SGD optimizer with batch size $512$ and learning rate $0.2$. We reported the separately evaluated results on subsets of 'many-shot' (each class has more than $100$ samples), 'medium-shot' (each class has more than $20$ but less than $100$ samples), and 'few-shot' (each class has less than $20$ samples). The number of overall training epochs is $200$.  As to the method "Deferred Re-weighting"~\cite{LDAM}, this baseline method up-weights the learning of the rare classes and down-weights the learning of the common classes to create a relatively balanced training distribution in the latter stages of training. We defer the training with this class-balanced re-weighting method after $180$ epochs of regular training. As to the method, "Decoupling Reps\&Cls (CRT)"~\cite{Kang2020Decoupling}, this baseline method first trains the model with the regular training, and then fix representations but only re-initializes and re-train the classifier with a re-sampling strategy that up-weights the learning of rare classes and down-weights the learning of common classes for additional $30$ epochs.

\subsubsection{OOD detection}
We can improve the performance of OOD detection since EL sharpens the energy surface around the data.
For OOD detection, we use an indicator to detect whether samples are in the distribution (In-D) of the training set or out of the distribution (OOD). We use the minimum conditional energy $min_{y' \in\mathbb{Y}} E_{\vtheta}(y' \mid \vx)$ as the indicator.
We conduct experiment on a small-scale OOD detection task (In-D: MNIST, OOD: Fashion-MNIST) and a large-scale task (In-D: ImageNet, OOD: iNautralist2018). Samples in Fashion-MNIST are clothes (10 calsses), while samples in MNIST are digital numbers (from 0 to 9); samples in ImageNet are instances of 1,000 objects, while samples in iNautralist2018 are instances of  8,142 species. Samples in Fashion-MNIST and MNIST are  Gray-scale pictures, but samples in ImageNet and iNautralist2018 are RGB pictures. MNIST and Fashion-MNIST  both have 60,000 training examples. ImageNet and  iNautralist2018 have 1.2M and  437,513 training examples respectively.  

\subsubsection{Adversarial Robustness}
In Section 4.4 of the main paper, to verify the claim that well-classified examples help improve robustness, we use FGSM 
~\cite{fgsm} and PGD~\cite{pgd} to construct adversarial samples to attack the models trained with CE loss and the models trained with encouraging loss, respectively.  In the Appendix, we also test robustness under Auto Attack~\cite{auto-attack}. We perform experiments on MNIST, the implementations  of these three methods and the base model are from TorchAttacks~\cite{kim2020torchattacks}. We adopt their settings for training and evaluation on MNIST since they get 99\% accuracy on MNIST with significantly less training time than ~\cite{l-softmax}, but ~\cite{l-softmax} get higher accuracy with 60x epochs. To report improvements on strong baselines, we take the setting of ~\cite{l-softmax} for experiments  in Table 1 and Table 5 in the main paper.  The short training epochs of TorchAttacks can save much time in adversarial training too since we generate inputs in every step. Thus, to be consistent with experiments in adversarial training, we adopt the setting from TorchAttacks.

\subsection{Experiment Setups for New Experiments in the Appendix}
\subsubsection{Object Detection}\label{sec:detection} 
For experiments on object detection, we perform experiments on the common-used COCO dataset~\footnote{COCO detection data: \url{https://cocodataset.org}}. 
we adopt the configuration of ``RetinaNet-R-50-FPN-1x'' from the github repository Detectron2\footnote{Environment of detection experiments:\url{https://github.com/facebookresearch/detectron2}} as our default setting. 
In this setting, the one-stage detector of RetinaNet~\cite{focal_loss} with the backbone ResNet50 is trained for $90k$ updates, and each batch has $8$ images.

\subsubsection{Adversarial Training}
In the Appendix, we demonstrate our idea is additive to Adversarial Training. We first train the model with the input images generated by adversarial attack methods iteratively and then test the model with corresponding attack methods. The implementation for adversarial training is from TorchAttack~\cite{kim2020torchattacks}.

\subsection{Training Costs}
\begin{table*}[t] 
    \centering
    \small
    \begin{tabular}{@{}l c r r r r l@{}} 
    \toprule
         Data&Infrastructure &Mem/GPU & Time & Epochs &Samples & Model \\ \midrule
         MNIST & RTX 2080Ti * 1  & 3G &0.3h & 100 &60K &CNN-Arch~\cite{l-softmax} \\
         CIFAR-10 & RTX 2080Ti * 1  & 7G &5h & 90 &50K &ResNet50 \\
                  CIFAR-10 & RTX 2080Ti * 1  & 5G &5h & 90 &50K &EfficientNet-B0 \\
         CIFAR-100 & RTX 2080Ti * 1  & 7G &5h & 90 &50K & ResNet50 \\
         CIFAR-100 & RTX 2080Ti * 1  & 5G &5h &90 &50K & EfficientNet-B0 \\
         ImageNet & TITAN RTX * 2 & 13G &48h &90 &1.2M  &ResNet50 \\
         ImageNet & TITAN RTX * 2 & 11G &158h &450 &1.2M  & EfficientNet-B0 \\
         PROTEINS & RTX 2080Ti * 1 & 0.7G &27s &85 &1113 &GCN+SAGPool\\
         NCI1 & RTX 2080Ti * 1 & 0.7G &260s &190 &4110 &GCN+SAGPool\\
         Fr-En translation  & RTX 2080Ti * 1  & 4G &4h  &-  & 233K & Transformer\\
         De-En translation  & RTX 2080Ti * 1  & 5G &4h  &- & 160K & Transformer \\
        iNaturalist 2018 & TITAN RTX * 4 & 15G &48h &200 &438K &ResNet50 \\
         COCO Detection & RTX 2080Ti * 4 &9G &7h &- &118K & ResNet50 +RetinaNet\\\bottomrule
    \end{tabular}
        \caption{Training costs of each setting. Samples are sentences pairs in translation, images in image classification tasks, and graphs in graph classification. The training steps are calculated in the level of updates on Fr-En translation, De-EN translation, and COCO detection. For Fr-En translation, we train the model for $100K$ steps and each batch has $4K$ tokens. For De-En translation, we train the model for  $20K$ steps and each batch has $16K$ tokens. For COCO detection, we train the model for $90K$ steps and each batch has $32$ images.}
        \label{tab:training_costs}
\end{table*} 

The training costs are summarized in Table \ref{tab:training_costs}, and we estimate that the total cost for this study is running a GPU of RTX 2080TI for 300 days.


\section{Supplementary Results}

\subsection{Well-classified examples are also underestimated in the improved variants of CE loss} \label{sec:var}
\subsubsection{Adversarial Training}
We also perform experiments on adversarial training with FGSM (Linf), PGD (Linf). We didn't do that for AutoAttack since it is too slow. Following the script in "implementation from "https://github.com/Harry24k/adversarial-attacks-pytorch/demos/Adversairal Training (MNIST).ipynb.", we generate adversarial training examples at every step and use them for training the model.
The results in Table \ref{tab:adv-training} show that EL is additive with adversarial training.
 We note that EL does not further benefit from adversarial training under the attack of L2-norm PGD. The reason may be that models trained with EL and clean data are already robust to L2-norm PGD attack.
 
\begin{table*}[]
\small
\centering
\begin{tabular}{llllllll}
 &FGSM(Linf)       &       & PGD(Linf) &       & PGD (L2)   &       &       \\ 
 \toprule
Epsilon    & CE    & EL    & CE        & EL    & Epsilon    & CE    & EL    \\  \midrule
No\_attack & 95.43 & 97.21 & 94.3      & 96.88 & No\_attack & 98.97 & 98.98 \\
0.05       & 92.59 & 95.81 & 92.32     & 95.77 & 0.3        & 98.38 & 98.4  \\
0.1        & 91.01 & 95.18 & 89.89     & 94.48 & 1          & 93.88 & 94.36 \\
0.15       & 89.79 & 95.16 & 87.07     & 93.03 & 3          & 10.83 & 76.22 \\
0.2        & 89.3  & 95.33 & 84.51     & 91.79 & 10         & 0.01  & 76.22 \\
0.25       & 89.09 & 95.09 & 82.17     & 90.84 & 30         & 0.01  & 76.22 \\
0.3        & 87.87 & 94.15 & 79.72     & 90.03 & 100        & 0.01  & 76.22 \\ \bottomrule
\end{tabular}
\caption{Adversarial robustness after adversarial training under the attack of FGSM and PGD (Linf and L2). For generating adversarial training examples, the epsilon is 0.3 for two Linf norm attack methods and 3 for the PGD with L2 norm restricted attacks.}
\label{tab:adv-training}
\end{table*}

\begin{figure}
    \centering
    \includegraphics[width=1.0\linewidth]{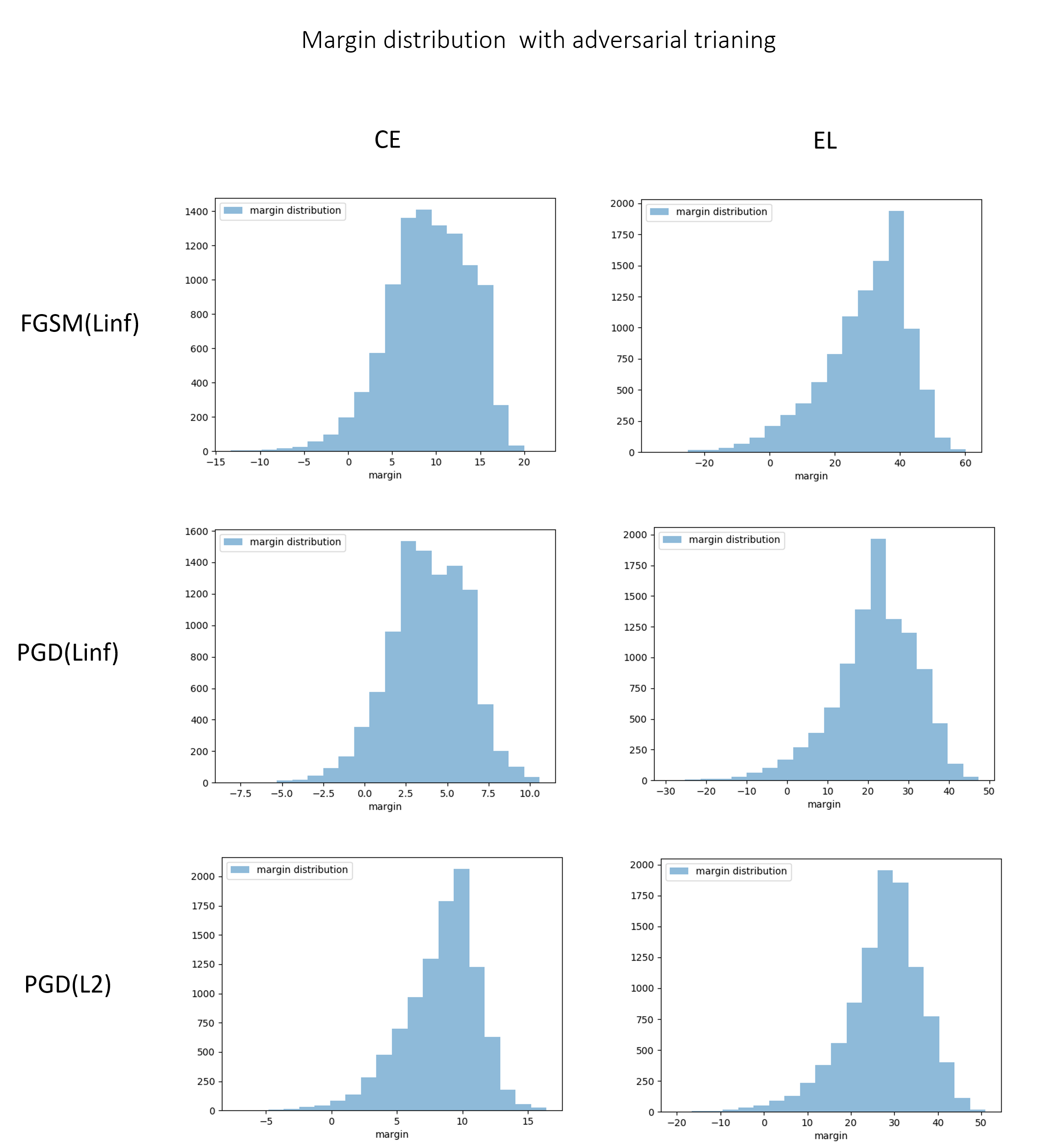}
    \caption{Margin distribution under adversarial training.}
    \label{fig:adv-margin}
\end{figure}
We plot margin distributions on MNIST in the above three adversarial scenarios and demonstrate EL can also enlarge margin in these scenarios. We can see from  Figure \ref{fig:adv-margin}  that EL increases the margin for most samples compared to CE, even with adversarial training.

\subsubsection{Large margin softmax}
On MNIST, we add encouraging loss with large margin softmax~\cite{l-softmax}, the accuracy is improved from 99.62$\pm$ 0.04 to 99.64 $\pm$ 0.02.  Since they do not provide code for other tasks and do not work for other tasks with the provided code.  The accuracy under attack is improved from 92.89 to 93.81 with epsilon(perturbation amount) 0.2, and 82.68 to 86.02 with epsilon 0.3. In addition, methods of increasing margins do not show the improvement on ImageNet, but we make it, as shown in the main paper.

\subsubsection{Label Smoothing}
The translation experiments are with label smoothing. We show that our method is also additive to pure CE loss, and the improvements are on par with label smoothing.  We show these results in Table \ref{tab:ls}.

\begin{table*}
  \caption{Well-classified examples also help improve CE loss with label smoothing, and the improvements are on par with it.}
  \centering
  \small
  \begin{tabular}{llll}
    \toprule
    \cmidrule(r){2-4}
    Name     & MNIST    & Fr-En Translation & De-En Translation \\
    \midrule
    CE loss &99.42 $\pm$ 0.06 &36.66 $\pm$ 0.14 &34.03 $\pm$ 0.06 \\
    + label smoothing & 99.54 $\pm$ 0.04  & 37.14 $\pm$ 0.15 &35.09 $\pm$ 0.07 \\
    Encouraging Loss &99.56 $\pm$ 0.05 &37.36 $\pm$ 0.08 & 34.60 $\pm$ 0.12 \\
    + label smoothing &99.63 $\pm$ 0.04 & 37.80 $\pm$ 0.17 & 35.50 $\pm$ 0.11\\
    \bottomrule
  \end{tabular}
  \label{tab:ls}
\end{table*}
\subsubsection{Focal loss}
Focal Loss~(FL)  \cite{focal_loss} propose that well-classified examples should be more overlooked than CE loss in imbalanced classification,  its form is:
\begin{equation}
    \mathcal{L} = -(1-\vp_y)^{{\gamma}_{f}} {\cdot} (\vy {\cdot} \log \vp), 
\end{equation}

\begin{figure*}[!h]
\begin{minipage}{.3\linewidth}
\centering
\includegraphics[width=1.1\linewidth]{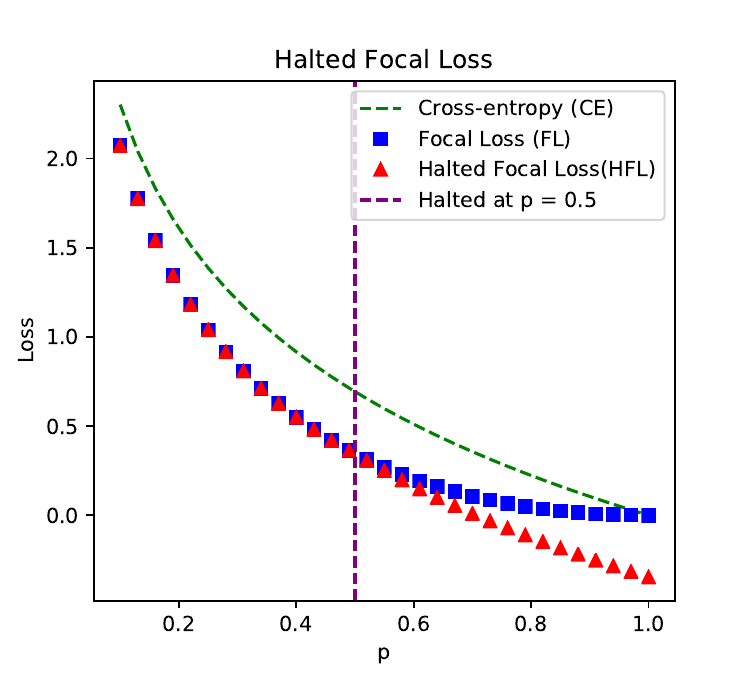}
\caption{Illustration of Halted Focal Loss which reverts Focal Loss to CE Loss when the correctness of a example is high.
}
\label{fig:bg}
\end{minipage}
\hfil
\begin{minipage}{.7\linewidth}
\centering

\scalebox{0.68}{
\begin{tabular}{lllll}
\toprule
\multirow{1}{*}{Method}  & \multicolumn{4}{c}{iNaturalList2018}                  \\
                                              & Overall & Many & Medium & Few \\ \midrule
Focal Loss~\cite{focal_loss}                      & 62.9 & 72.9 & 64.1 &58.8             \\ 
Halted Focal Loss                     & 63.6(+0.7)    & 73.1(+0.2)     &  64.6(+0.5)      & 59.7(+0.9)                \\
\midrule
Cross Entropy Loss & 64.3 &74.1 &65.9 &59.8               \\ 
Encouraging Loss (conservative)  &65.3 &74.3 &66.6 &61.2 \\ 
Encouraging Loss  & 65.8 &74.4 &66.6 &62.4 \\
Encouraging Loss (aggressive) & 66.3 (+2.0) & \textbf{75.1} (+1.0)     & 67.4(+1.6)       & 62.6(+2.8) \\
\bottomrule
\end{tabular}
}
\makeatletter\def\@captype{table}\makeatother 
\caption{Comparison between Focal Loss, Halted Focal Loss, CE Loss and  Encouraging Loss on long-tailed classification dataset iNaturalist 2018, std is about 0.4.}
\label{tab:imb-focal}
\end{minipage}
\end{figure*}

We propose a variant of the focal loss, the Halted Focal Loss (HFL), such that the high-likelihood area is not prioritized. 
The HFL reverts the focal loss to the cross-entropy Loss when the likelihood is high enough:
\begin{equation}
    \mathcal{L} = \left\{ \begin{array}{ll} -(1-p_y)^{{\gamma}_{f}} {\cdot} (\vy {\cdot} \log \vp), & \text{if } p_y \leq \varphi \\ - \vy{\cdot}[\log \vp + b], & \text{otherwise} \end{array} \right.
\end{equation}
Where $p_y$ is the prediction probability of the correct label, $ b = \alpha(1- (1-\varphi)^{{\gamma}_f}) \log \varphi$ to ensure monotonicity and continuity, and $\varphi$ is the boundary between the low-likelihood and the high-likelihood area, which we set as $0.5$, i.e., a likelihood higher than which renders a definitely well-classified example. This halted focal loss is plotted in Figure~\ref{fig:bg}, which has the same gradients as the cross-entropy in the high-likelihood area and remains the same as the Focal Loss in the low-likelihood area. 

On imbalanced classification, we find that the correct predictions matter for improving accuracy.  Results are shown in Table \ref{tab:imb-focal} in terms of experiments on iNaturalist 2018 and in Table \ref{tab:detection} on COCO detection.

\begin{table}
  \small
  \centering
\setlength{\tabcolsep}{5pt}
  \begin{tabular}{@{}llll@{}}
    \toprule
    Method      & $AP$   & $AP_{50}$   & $AP_{75}$    \\
    \midrule
    FL & 34.45 $\pm$ 0.13 & 54.23 $\pm$ 0.18 &36.66 $\pm$0.13  \\  %
    HFL  &34.60 $\pm$ 0.09  & 54.24 $\pm$ 0.15 &36.73 $\pm$ 0.18 \\  
    \bottomrule
  \end{tabular}

  \makeatletter\def\@captype{table}\makeatother
  \caption{Results of HFL (Halted Focal Loss) on the COCO detection dataset,
  AP denotes average precision.As we can see, increasing the importance of the well-classified examples with halted focal loss achieves better results.
  }
    \label{tab:detection}
\end{table}

\subsection{Other Results} \label{sec:other-res}
\subsubsection{EL Reduces Excess Risk}
EL reduces the gap between train acc and test acc by 2 points, 3 points, and 2.5 points for training ResNet50 on CIFAR-10, CIFAR-100, ImageNet, respectively.

\subsubsection{EL with Conservative Bonus also Reduces Energy on the Data and Enlarges Margins} 
\begin{figure}
    \centering
    \includegraphics[width=1.0\linewidth]{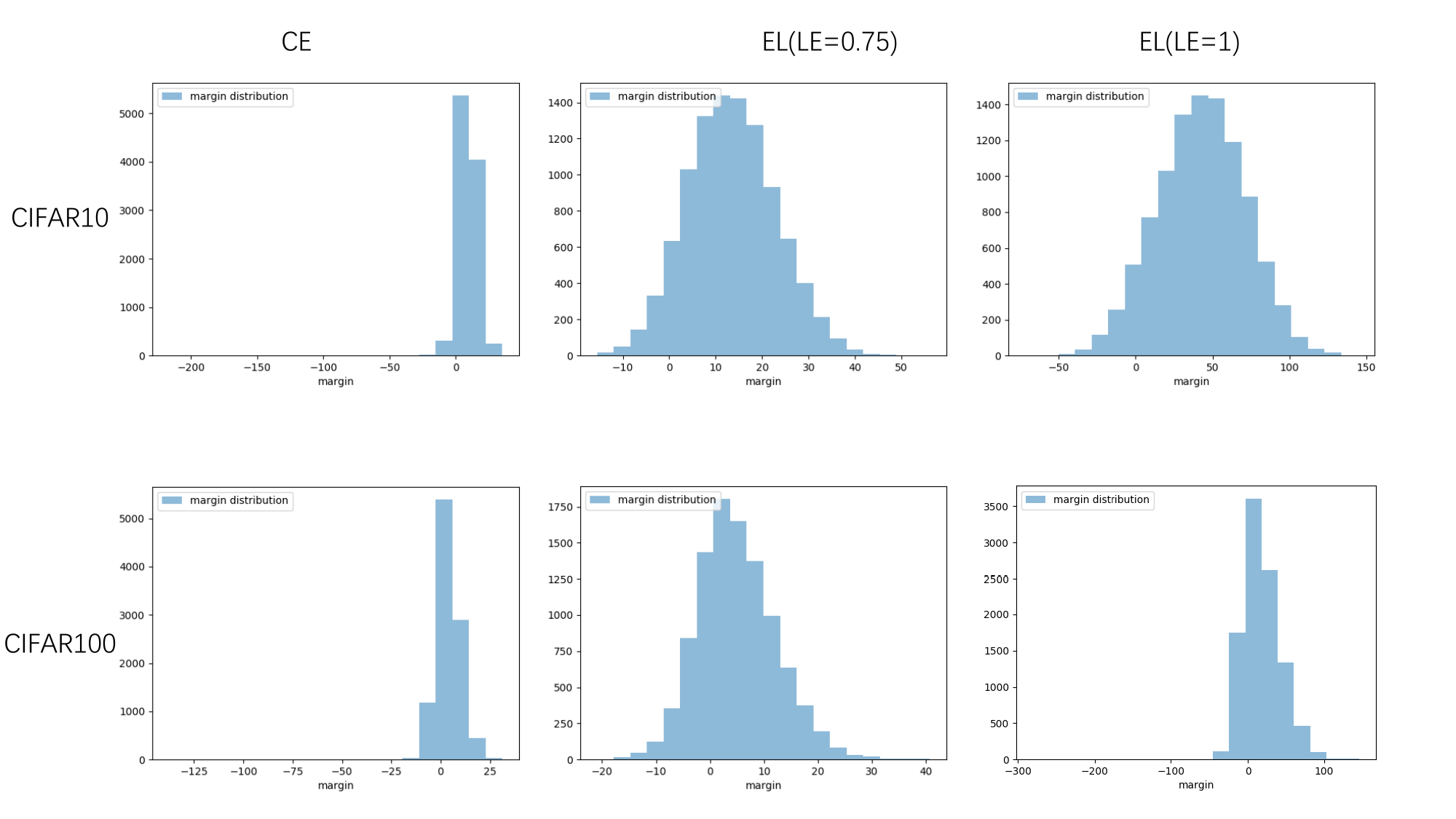}
    \caption{EL enlarges margins on CIFAR-10 and CIFAR-100.}
    \label{fig:margin-c10100}
\end{figure}
For a conservative bonus, the margin is from -60 to 60, centered by 17. And the energy for the target class "cat" is -17, which is -5 lower than that of CE loss.
We plot  margin distribution on CIFAR-10 and CIFAR-100 in Figure \ref{fig:margin-c10100}

\subsubsection{Adversarial Robustness under AutoAttack}
Beyond the canonical attack methods  PGD, we also conduct experiments with  a recently proposed method, AutoAttack~\cite{auto-attack}.  The results in Table \ref{tab:auto-attack} show that models trained with EL are also more robust to  AutoAttack. 
 \begin{table}[]
\begin{tabular}{llllll} \toprule
MNIST      & \multicolumn{2}{l}{AutoAttack   (Linf)} & \multicolumn{2}{l}{AutoAttack  (L2)} &       \\ 
Epislon    & CE                 & EL                 & Epislon             & CE             & EL    \\ \hline
No\_attack & 98.69              & 99.04              & No\_attack          & 98.69          & 99.04 \\
0.05       & 93.51              & 94.99              & 0.3                 & 96.76          & 97.54 \\
0.1        & 74.89              & 79.31              & 1                   & 80.76          & 84.03 \\
0.15       & 32.98              & 37.74              & 3                   & 0.05           & 0.27  \\
0.2        & 5.04               & 7.59               & 10                  & 0              & 0     \\
0.25       & 0.01               & 0.26               & 30                  & 0              & 0     \\
0.3        & 0                  & 0                  & 100                 & 0              & 0    \\ \bottomrule
\end{tabular}
\caption{Adversarial robustness under the attack of AutoAttack.}
\label{tab:auto-attack}
\end{table}

\begin{table}[]
\small
\begin{tabular}{l|ll} \toprule
  Setting & \multicolumn{2}{l}{MNIST vs. FashionMNIST}  \\ 
Metric& AUROC$\uparrow$                 & FPR95$\downarrow$        \\ \midrule
CE & 95.68                 & 20.92                \\
CE+label smoothing &97.10 &	8.10\\
EL & 98.04                 & 8.69            \\ 
EL+label smoothing &99.55&	0.98 \\
\bottomrule   

\end{tabular}
\caption{EL is additive with label smoothing  for improving OOD detection performance. Higher AUROC and lower FPR95 are better.}
\label{tab:ood-ls}
\end{table}

 \begin{table*}[]
\centering
\small
\begin{tabular}{l|lll|lll} \toprule
                                              & \multicolumn{3}{c}{w/o LS}                                                                                          & \multicolumn{3}{c}{w LS}                                                                                             \\
\multicolumn{1}{c}{\multirow{2}{*}{CIFAR-10}} & \multicolumn{1}{c}{\multirow{2}{*}{ECE}} & \multicolumn{2}{c}{Energy}                                               & \multicolumn{1}{c}{\multirow{2}{*}{ECE}} & \multicolumn{2}{c}{Energy}                                                 \\
\multicolumn{1}{c}{}                          & \multicolumn{1}{c}{}                     & \multicolumn{1}{c}{on the data} & \multicolumn{1}{c}{lowest out-of-data} & \multicolumn{1}{c}{}                     & \multicolumn{1}{c}{on the data} & \multicolumn{1}{c}{lowest out-of-data}  \\ \midrule
CE                                            & 0.031                                    & -15.1                           & -6.2                                   & 0.088                                    & -5                              & -2.3                                   \\
EL(LE=0.75)                                   & 0.042                                    & -20.8                           & -7.4                                   & 0.034                                    & -4.8                            & -1.2                                   \\
EL(LE=1)                                      & 0.062                                    & -54.5                           & -11.73                                 & 0.056                                    & -12.3                           & -0.8                                    \\ \midrule
                                              & \multicolumn{3}{c}{w/o LS}                                                                                          & \multicolumn{3}{c}{w LS}                                                                                           \\ 
\multicolumn{1}{c}{\multirow{2}{*}{MNIST}}    & \multicolumn{1}{c}{\multirow{2}{*}{ECE}} & \multicolumn{2}{c}{Energy}                                               & \multicolumn{1}{c}{\multirow{2}{*}{ECE}} & \multicolumn{2}{c}{Energy}             \\
\multicolumn{1}{c}{}                          & \multicolumn{1}{c}{}                     & \multicolumn{1}{c}{on the data} & \multicolumn{1}{c}{lowest out-of-data} & \multicolumn{1}{c}{}                     & \multicolumn{1}{c}{on the data} & \multicolumn{1}{c}{lowest out-of-data}  \\ \midrule
CE                                            & 0.005                                    & -27                             & -5.85                                  & 0.009                                    & -4                              & -0.08                                   \\
EL(LE=1)                                      & 0.004                                    & -131.6                          & -23.9                                  & 0.003                                    & -12.6                           & -0.07                                 
\end{tabular}
\caption{We compared the ECE and Energy of EL and CE with and without label smoothing. Label smoothing reduces ECE when using EL.}
\label{tab:ls-ece}
\end{table*}

\subsubsection{Maximum Conditional Energy is a Good Indicator for OOD Detection}
\begin{figure}
    \centering
    \includegraphics[width=1.0\linewidth]{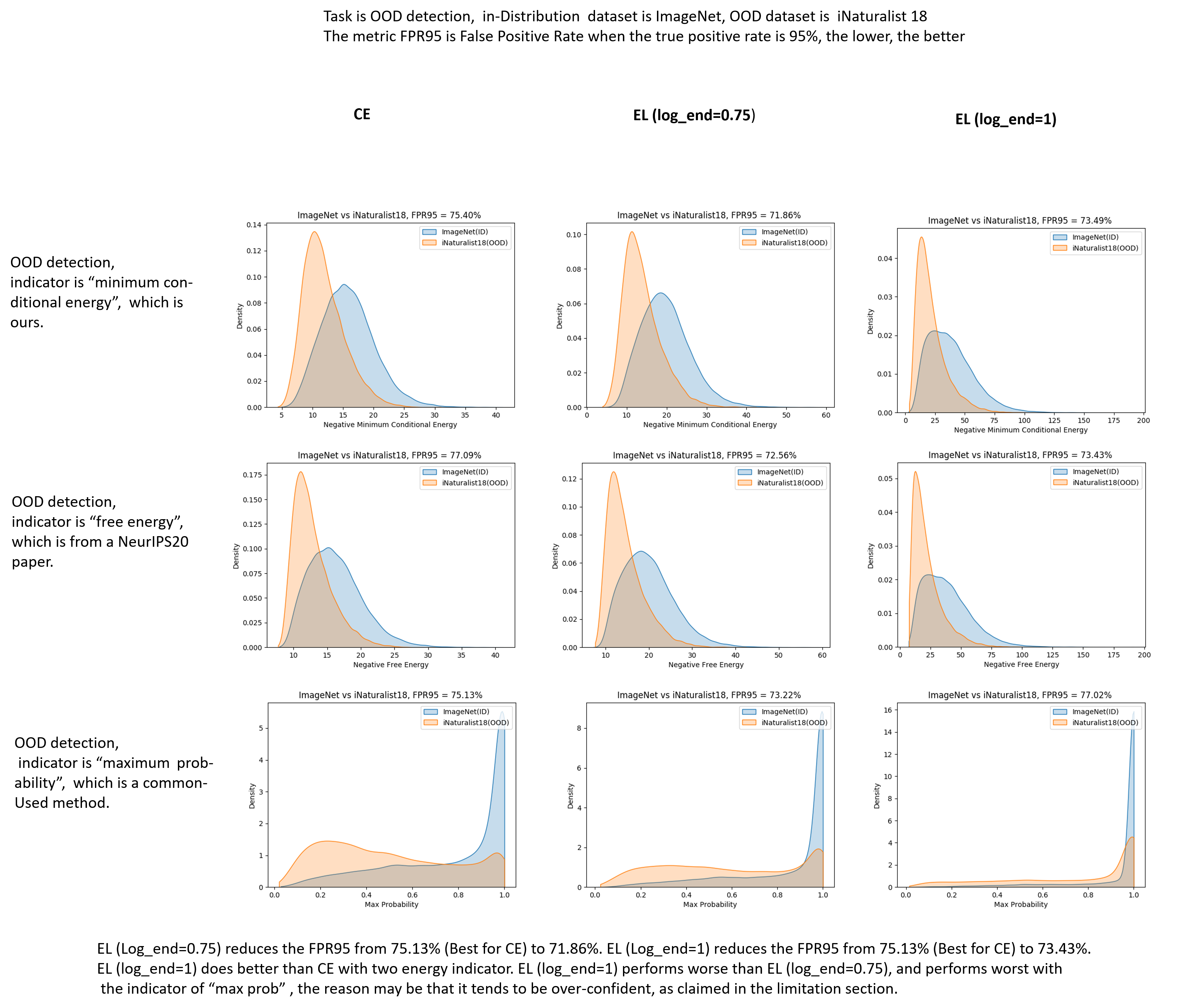}
    \caption{OOD detection on ImageNet vs. iNautralist2018}
    \label{fig:ood-imagenet}
\end{figure}

We do experiments on a large-scale OOD detection task, in which In-D data is ImageNet(It contains 1.2M images, including 50K images for the test), and OOD data is iNautralist2018 (It contains 438 K samples, including 24.4K images for the test). At the same time, MNIST only has 70k low-resolution images in total. ImageNet contains 1K common objects, while iNautralist2018 contains 8,142 species. These datasets are large so that the OOD detection is challenging. 
We show the results in Figure \ref{fig:ood-imagenet}.
Our indicator “minimum conditional energy among all classes,  The “free energy”~\cite{free-energy} method uses logarithm sum exp of energy among all classes. The difference is that “free energy” considers overall energy. The similarity between us and them is that we all consider energy, but the difference is that we consider the maximum value among classes, which is similar to ”max prob.”

We can also learn from figures that with all these three indicators, EL (LE=0.75) does better for OOD detection than CE, we reduce the FPR95 from 75.13\% (Best for CE) to 71.86\%, and EL (LE=1) does better than CE with two energy indicator. These results demonstrate that the EL improves the energy distribution in the case of OOD detection.

\subsubsection{Label Smoothing Helps reduce  ECE in EL}
We in Table \ref{tab:ls-ece} that Label Smoothing (LS) can reduce ECE for EL. Although label smoothing reduces the energy values for both CE and EL, we can see that in EL, the difference in energy value between non-data points and data points is still more significant than in CE. For example, on MNIST, after applying label smoothing, the average energy of EL at data points is lower than CE (EL: -12.6, CE:-4), while the average energy of non-data points is slightly higher than CE (EL: -0.07, CE:-0.08), on CIFAR-10, we can push down the energy on the data form -5 to -12.3 while pushing up the lowest energy out of the data from -2.3 to -0.8.
\subsubsection{The combination of EL and label smoothing can significantly improve OOD detection}
We show in Table \ref{tab:ls-ece} that EL is much better than CE, even with LS. For example, we reduce the metric FPR95 from 8.10\% to 0.95\% on MNIST vs. Fashion MNIST. This is because we get a more discriminative energy distribution.


\section{Other Related Work}
\paragraph{Rewarding the model}
 The idea of encouraging loss that rewarding correct predictions is partly inspired by Reinforcement Learning (RL)~\cite{sutton1998introduction,williams1992simple}. However, we are different since we take the likelihood into consideration when designing loss, and our goal is to make correct predictions take a more important role than before during learning.
 considering likelihood in design rewards and let correct predictions take a more important role in learning.
 
\noindent \paragraph{Discussion About Mixup}
Since the augmentation method Mixup~\cite{zhang2018mixup} that interpolates both input and output of two examples, unexpectedly pays attention to well-classified examples. So the idea cannot further improve it. However, the application of Mixup is limited because of its bad understandable semantics. 
including AugMix~\cite{augmix} and Token labeling~\cite{jiang2021token} can be combined with this technique. 
Moreover, the SoTA parameter-efficient model Efficientnet-B0 does not use the Mixup but uses the RandAugment~\cite{rand-aug}, our idea improves it.
Mixup improves the Resnet50 by 0.2 with the same number of epochs (90 epochs) and improves by 1.5 with more than 2x epochs, while we can improve the ResNet-50 by 0.9 without using extra epochs.

\noindent \paragraph{Poly Loss} To build loss functions that are flexible to various tasks and datasets, Poly Loss\cite{leng2022polyloss} design a group of loss functions that can be a linear combination of polynomial functions, and reached good performance on multiple image classification data. Our theory can explain the success of poly loss, and the polynomial additional term of poly loss also enhances the learning weight of well-classified samples. Their optimal choice is similar to the case of LE=0 in Encouraging Loss. We found that the key is not meet a polynomial form, but the role of well-classified samples in learning. Therefore, we designed a more optimal situation, such as Encouraging Loss (LE=0.75).
\end{document}